\documentclass[runningheads]{llncs}

\usepackage{eccv}

\usepackage{eccvabbrv}

\usepackage{graphicx}
\usepackage{booktabs}

\usepackage[accsupp]{axessibility}  

\usepackage{hyperref}

\usepackage{orcidlink}

\begin{document}

\title{Boosting 3D Foundation Models with \\ Edge-based Pose Optimization} 

\titlerunning{Boosting 3D Foundation Models with Edge-based Pose Optimization}

\author{
    Mattia D'Urso\inst{1}\orcidlink{0009-0003-3672-6559} \quad 
    Christian Sormann\inst{2}\orcidlink{0000-0002-6824-4007} \quad 
    Mattia Rossi\inst{2}\orcidlink{0000-0001-5158-2395} \quad  \\
    Friedrich Fraundorfer\inst{1}\orcidlink{0000-0002-5805-8892}
}

\authorrunning{M.~D'Urso et al.}

\institute{$^{1}$Graz University of Technology \quad $^{2}$Sony Europe}

\maketitle
\begin{figure*}[h]
    \renewcommand\twocolumn[1][]{#1}%
    \centering
    \setlength{\fboxrule}{0.8pt} 
    \setlength{\fboxsep}{0pt}    

    \begin{subfigure}[t]{0.325\textwidth}
        \centering
        \fcolorbox{black}{white}{\includegraphics[width=\textwidth]{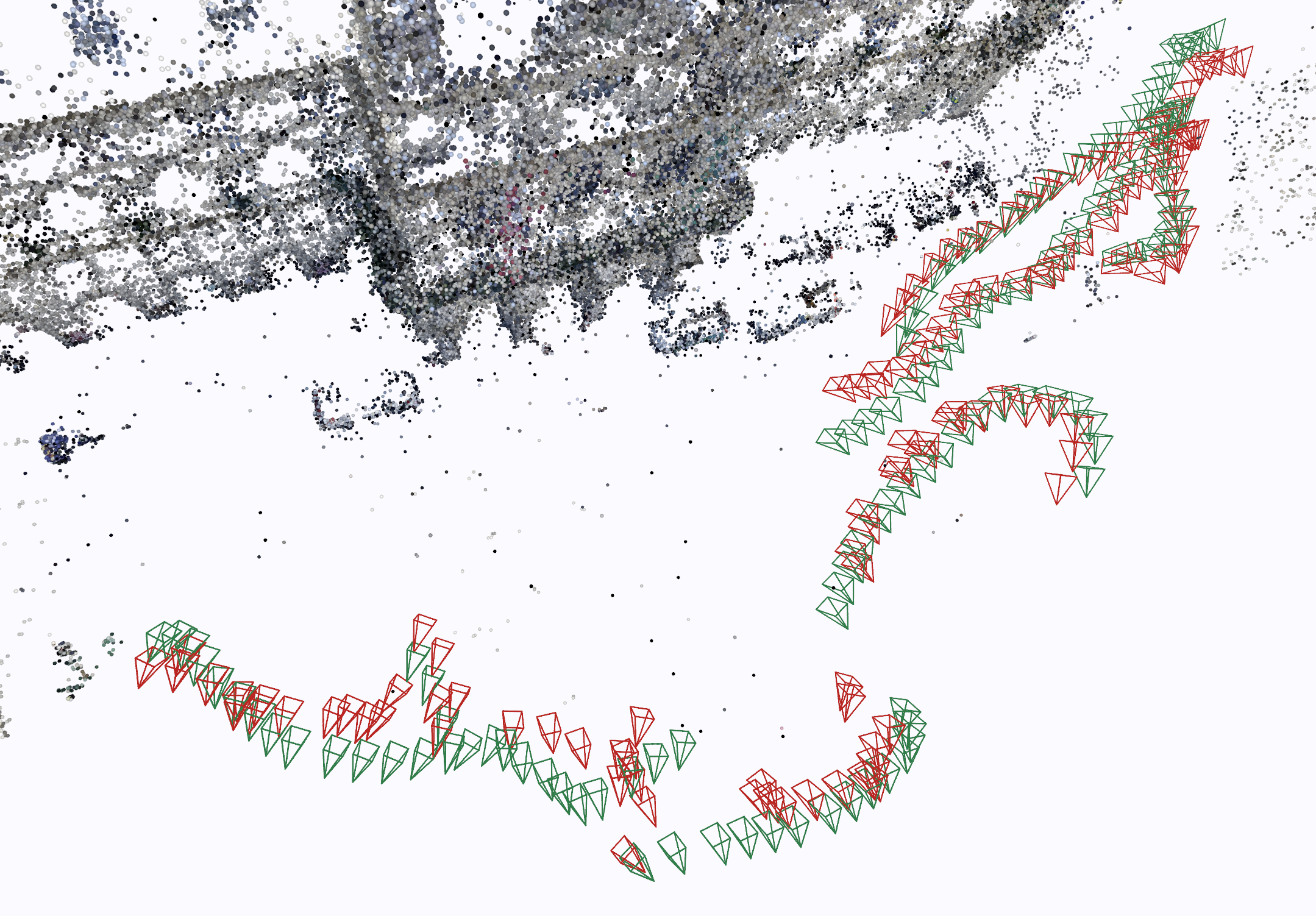}}
        \caption{Initial VGGT output}
        \label{fig:rerun_0}
    \end{subfigure}
    \hfill 
    \begin{subfigure}[t]{0.325\textwidth}
        \centering
        \fcolorbox{black}{white}{\includegraphics[width=\textwidth]{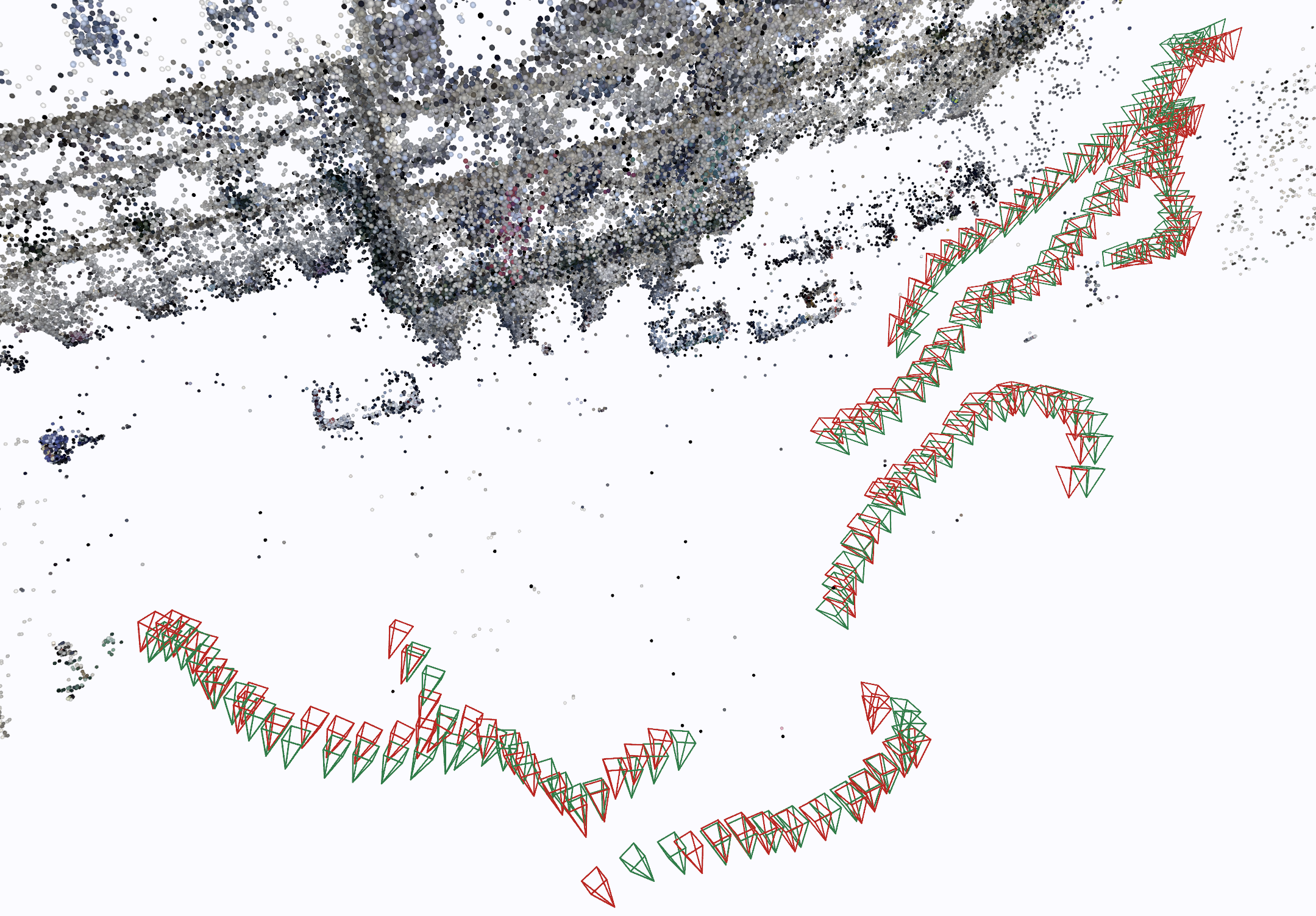}}
        \caption{Intermediate step}
        \label{fig:rerun_1}
    \end{subfigure}
    \hfill
    \begin{subfigure}[t]{0.325\textwidth}
        \centering
        \fcolorbox{black}{white}{\includegraphics[width=\textwidth]{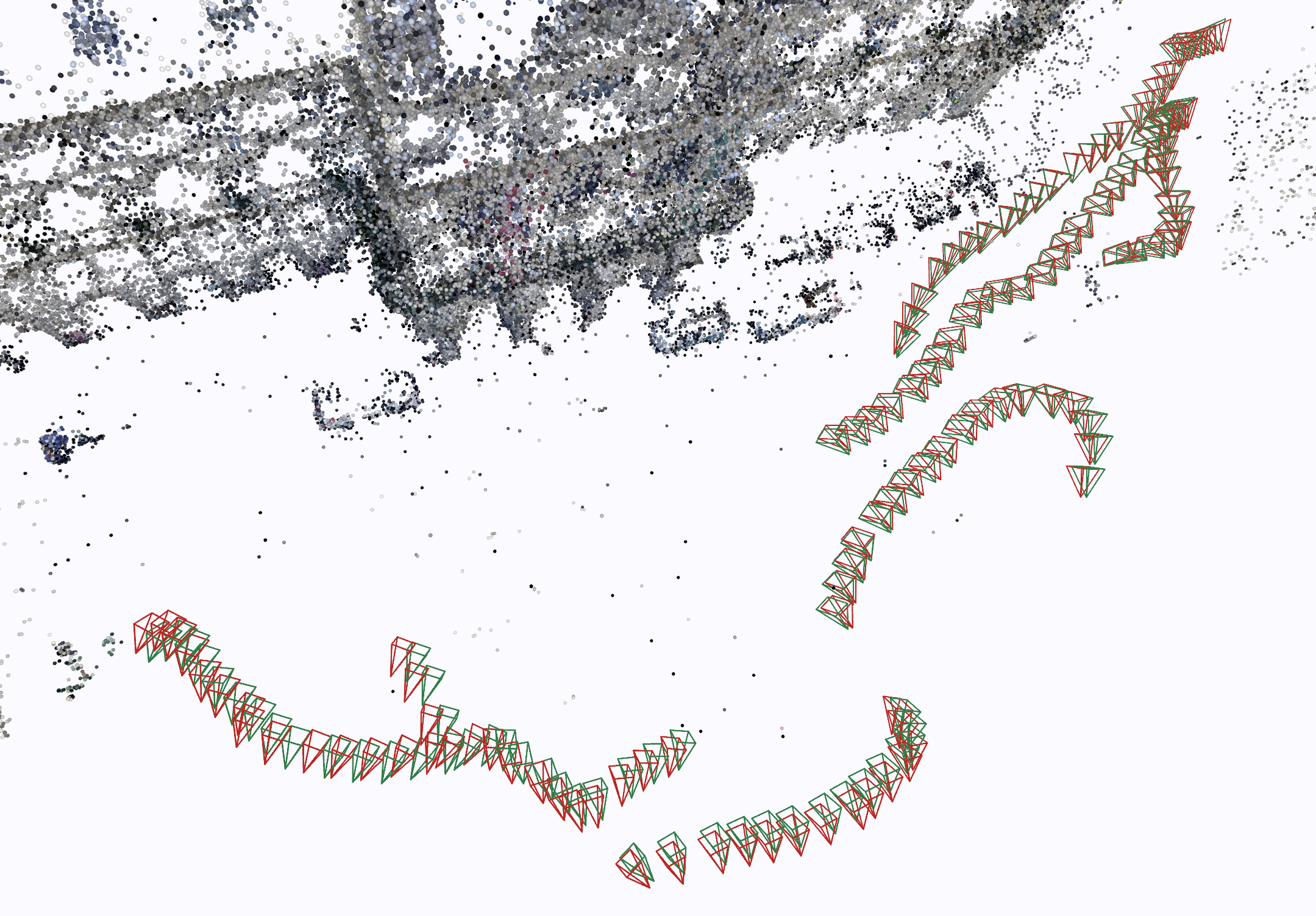}}
        \caption{Final convergence}
        \label{fig:rerun_3}
    \end{subfigure}

    \caption{Visualization of three stages of our pose optimization method and the ground truth sparse point cloud for the Graz Town Hall scene (TerraSky3D). Starting from the initial optimization stage (a), provided by the VGGT output, we illustrate an intermediate state (b) and the final refined poses (c). The ground truth poses are shown in {\color{RLGreen}green} and the optimized ones in {\color{RLRed}red}. 
    }
    \label{fig:rerun}
\end{figure*}



\begin{abstract}
We introduce \textbf{Edge-based Pose Optimization (EPO)}, a trackless geometric
optimization framework specifically designed to boost the Structure-from-Motion reconstructions
generated by 3D Foundation Models.
These models achieve rapid inference by bypassing the time-consuming feature extraction
and matching stages of traditional pipelines, where explicit correspondences between
each 3D point and multiple images, referred to as tracks, are established.
However, their geometric accuracy currently falls short of traditional pipelines.
While this can be addressed in a post-processing step via Bundle Adjustment-like
refinement, doing so requires extracting feature tracks, thus defeating the original
speed advantage.
Instead, our fully differentiable framework uses edge map alignment as a proxy
for geometric optimization, avoiding feature extraction and track construction entirely.
Through extensive evaluation across multiple datasets and tasks, we demonstrate that
EPO matches or outperforms Bundle Adjustment-like methods while requiring significantly
lower runtime and memory.
Notably, its reduced memory footprint makes EPO suitable for consumer-grade hardware,
where competing refinement methods cannot run. Code is available at \url{https://github.com/mattiadurso/EPO}.\\


\end{abstract}



\clearpage
\section{Introduction}
\label{sec:intro}

The problem of recovering the 3D structure of a scene  from a collection of 2D images, known as Structure-from-Motion (SfM), is a fundamental and longstanding problem in computer vision.  While the geometric foundations were established early on, in recent years both incremental \cite{schonberger2016structure, moulon2016openmvg, snavely2008modeling} and global \cite{pan2024global,sweeney2015theia, li2025fastmap, zhong2025instantsfm} pipelines have advanced in terms of accuracy, reliability, and scalability. These frameworks typically rely on feature extraction and matching to then estimate camera poses and 3D scene points jointly. These are then refined via Bundle Adjustment (BA) \cite{triggs1999bundle}, a non-linear optimization technique, normally leveraging the Levenberg-Marquardt solver, that minimizes the error between the observations of each reconstructed 3D point at the different images and its actual projections, referred to as reprojection error. Despite its robustness, traditional SfM often struggles with textureless environments or sparse-view scenarios, where point-based features \cite{lowe2004distinctive} are frequently unreliable.

Recent research has shifted toward replacing complex, handcrafted pipelines with
streamlined neural networks~\cite{wang2024vggsfm, wang2025vggt, keetha2025mapanything,
wang2025pi}, referred to as 3D Foundation Models (3DFMs). By leveraging vast quantities
of training data, these models generate reconstructions in a rapid, feed-forward fashion.
However, while 3DFMs produce results in seconds, their geometric precision rarely matches
that of traditional SfM pipelines. Although their estimates can be refined through
post-processing, conventional methods such as BA rely on point
features and tracks, which defeats the inherent speed advantage of these models.
Furthermore, existing global alignment frameworks, primarily designed for DUSt3R-like
models~\cite{wang2024dust3r, duisterhof2025mast3r} or long-sequence
scaling~\cite{yang2025fast3r, elflein2025light3r}, typically require explicit 3D point
clouds and server-grade hardware.

To address these limitations, we introduce \textbf{EPO}, a trackless refinement
framework that decouples pose optimization from explicit feature matching, thereby
preserving the speed advantages of feed-forward models. By avoiding the overhead
of track generation, EPO runs in seconds while maintaining geometric precision
comparable to BA-like refinement methods, as shown in~\cref{fig:rerun}. Our main
contributions are summarized as follows:

\begin{itemize}
    \item We propose a robust, differentiable, and trackless optimization framework
    that boosts the geometric accuracy of 3DFMs through neural pose refinement and
    first-order optimization.

    \item We introduce an edge-based reprojection loss and a pose-based early stopping
    criterion that together reduce runtime to few seconds compared to minutes needed by BA-based refinement methods, while requiring lower memory.

    \item We provide a comprehensive evaluation across diverse 3DFM, benchmarks and real-world
    scenarios, including TerraSky3D\cite{durso2026terrasky}, ScanNet++\cite{yeshwanth2023scannet++, lin2025depth}, and Mip-NeRF~360\cite{barron2022mip}, demonstrating that
    EPO consistently performs on par with, or superior to, state-of-the-art refinement
    methods.
\end{itemize}

\section{Related Work}
\label{sec:related}

\subsubsection{Geometric Priors and Optimization}
While sparse keypoints~\cite{lowe2004distinctive} have remained the \textit{de facto}
standard for establishing geometric correspondences in SfM, they often prove insufficient
in textureless or repetitive environments. Consequently, several works have explored
alternative geometric primitives. Research into line-based SfM~\cite{pautrat2023deeplsd,
pautrat2023gluestick, liu20233d} demonstrates that lines and segments provide superior
constraints in man-made environments, where point features are sparse or collinear.
Works on edge-map-based SLAM~\cite{schenk2017robust, schenk2019reslam, zhao2023edgevo,
zhou2018canny, kuse2016robust} have shown that edges, similarly to lines \cite{liu20233d, liu2024robust}, serve as
highly informative observations, preserving structural integrity in scenes where
point-based descriptors show poor performance. However, SLAM methods can exploit temporal continuity, incremental spatial priors, and often metrically accurate depth measurements to guide edge-based alignment,
whereas our setting involves unordered image collections with none of these
advantages, initialized only from noisy model-predicted depth, making trackless
geometric optimization inherently more challenging.
Building upon this paradigm, our work leverages 2D edge maps~\cite{canny2009computational}
and distance transform fields~\cite{felzenszwalb2012distance} to formulate a
differentiable objective function, enabling joint optimization of camera parameters
and semi-dense depth by minimizing the distance between projected edges and their 2D
observations, achieving robust alignment without explicit feature points or descriptors.

\subsubsection{Deep Learning for Tracking and Matching}
The evolution of SfM has seen handcrafted components systematically replaced by neural
networks. Initial efforts focused on learned feature extractors~\cite{detone2018superpoint,
dusmanu2019d2, tyszkiewicz2020disk, santellani2022md, revaud2019r2d2, zhao2023aliked, durso2026sandesc}
and learned matchers~\cite{sun2021loftr, sarlin2020superglue, lindenberger2023lightglue,
edstedt2024roma, edstedt2023dkm}. To address temporal consistency, sparse matching has evolved into
persistent point tracking~\cite{harley2022particle, doersch2023tapir}. Architectures such
as CoTracker~\cite{karaev2024cotracker, karaev2025cotracker3} track points jointly
across long sequences, effectively handling occlusions and maintaining trajectory
coherence. However, these models remain computationally demanding, often requiring
significant GPU memory for extended sequences or large keypoint budgets.

\subsubsection{End-to-End Pipelines and Foundation Models}
Recently, the field has transitioned toward ``all-in-one'' learned frameworks that infer 3D
attributes directly from image collections. VGGSfM~\cite{wang2024vggsfm} pioneered
this shift with a fully differentiable pipeline that recovers camera poses and 3D
structure by leveraging deep 2D point tracks and a differentiable BA layer. This
paradigm was significantly advanced by 3DFMs such as DUSt3R~\cite{wang2024dust3r},
which predicts, given a pair of images, the corresponding depth maps that can be used
to estimate camera intrinsics and extrinsics. This approach was further extended by
MASt3R~\cite{leroy2024grounding}, which augments the DUSt3R architecture with a
dedicated matching head. By regressing dense local features and employing a
computationally efficient reciprocal matching scheme, MASt3R achieves high-precision
correspondences while maintaining robustness to extreme viewpoint changes.
MASt3R-SfM~\cite{duisterhof2025mast3r} then scales this local feature matching
approach into a complete SfM solution, by first employing an efficient image retrieval
strategy to construct a viewgraph and then executing a global optimization scheme
refined through successive 3D and 2D loss functions.

Building on these advancements, VGGT~\cite{wang2025vggt} employs a transformer-based
architecture to infer camera poses, depth, and point tracks in a single feed-forward
pass. Subsequent research has significantly broadened this paradigm. For instance,
MapAnything~\cite{keetha2025mapanything} facilitates metric-scale reconstruction by
integrating diverse geometric and semantic priors directly into the inference process.
Complementing these efforts, Pi3~\cite{wang2025pi} introduces a
permutation-equivariant framework designed to ensure robust global registration across heterogeneous viewing conditions. Fast3R~\cite{yang2025fast3r} and Light3R~\cite{elflein2025light3r} further extend this paradigm to long-sequence settings, enabling efficient reconstruction from hundreds to thousands of images, while Spann3R~\cite{wang2024spann3r}, CUT3R~\cite{wang2025cut3r}, and MUSt3R~\cite{cabon2025must3r} pursue scalable and streaming reconstruction by predicting pointmaps in a shared coordinate frame without explicit global alignment.

Despite their impressive performance, modern 3DFMs often face a
trade-off between inference speed and geometric fidelity. Higher precision typically
requires BA-like refinement methods, which rely on feature-based tracks. Unfortunately,
building tracks as a post-processing step significantly increases reconstruction
runtime, defeating the speed advantage of feed-forward models. In this work, we propose
EPO, a trackless geometric optimization framework that preserves this speed advantage
while achieving geometric precision on par with, or superior to, BA-like refinement
methods.
\section{Preliminaries}
\label{sec:preliminaries}
3DFMs, such as VGGT~\cite{wang2025vggt}, employ a unified transformer-based
architecture that leverages Alternating-Attention layers to reconstruct 3D scene
parameters from a collection of $N$ input images $\{I_i\}_{i=1}^N$. The process
begins by patchifying the input images into visual tokens using a DINO-based
backbone~\cite{oquab2023dinov2}. These image tokens are augmented with learnable
camera and register tokens before being processed by a transformer backbone to
aggregate spatial and temporal context. The resulting tokens are then routed to
specialized prediction heads:

\begin{itemize}
    \item \textbf{Camera Head:} Uses self-attention followed by a linear layer to
    estimate the camera intrinsics $\mathbf{K}$ and extrinsic matrices
    $\mathbf{P} = [\mathbf{R} \mid \mathbf{t}]$, where $\mathbf{R}$ and $\mathbf{t}$
    denote rotation and translation, respectively.

    \item \textbf{Dense Head:} Uses a Dense Prediction Transformer~\cite{ranftl2021vision}
    to simultaneously predict per-pixel depth maps $\mathbf{Z}$, 3D point maps, and
    dense tracking features. The depth maps are subsequently unprojected to 3D to
    form a point cloud.

    \item \textbf{Tracking Head:} Uses a CoTracker-based
    architecture~\cite{karaev2024cotracker, karaev2025cotracker3} to establish
    consistent 2D correspondences across an unordered image set. While this module
    provides the geometric constraints required for BA-based scene refinement,
    achieving high 2D localization precision demands iterative refinement through
    dense feature correlation and self-attention, making this step computationally
    expensive and memory-intensive.
\end{itemize}

\section{Method}
\label{sec:method}
\begin{figure}[t]
    \centering
    \begin{minipage}[b]{0.40\textwidth}
        \centering
        \includegraphics[height=2.85cm]{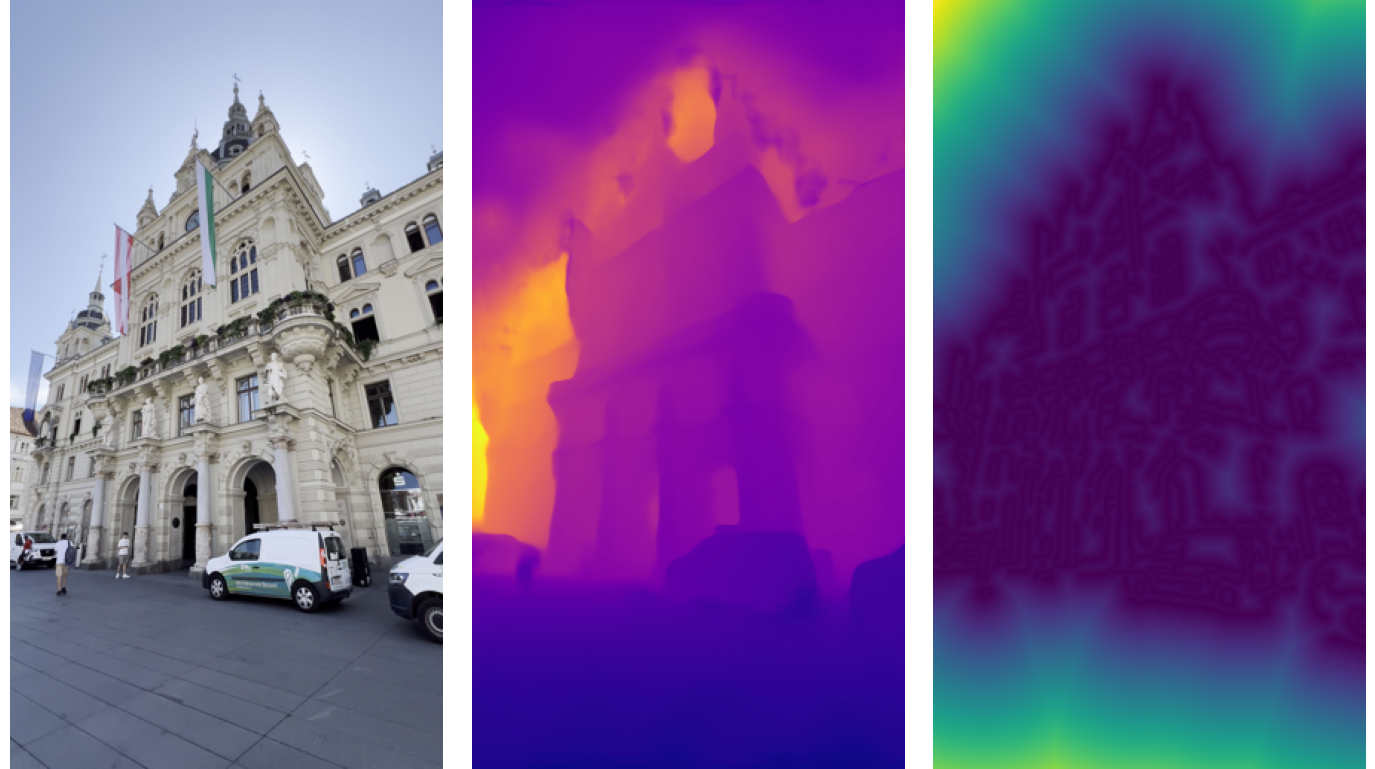}
        \caption{Example of input RGB image, VGGT raw depth, and the corresponding
        DTF.}
        \label{fig:rgb_data}
    \end{minipage}
    \hfill
    \begin{minipage}[b]{0.56\textwidth}
        \centering
        \includegraphics[height=2.85cm]{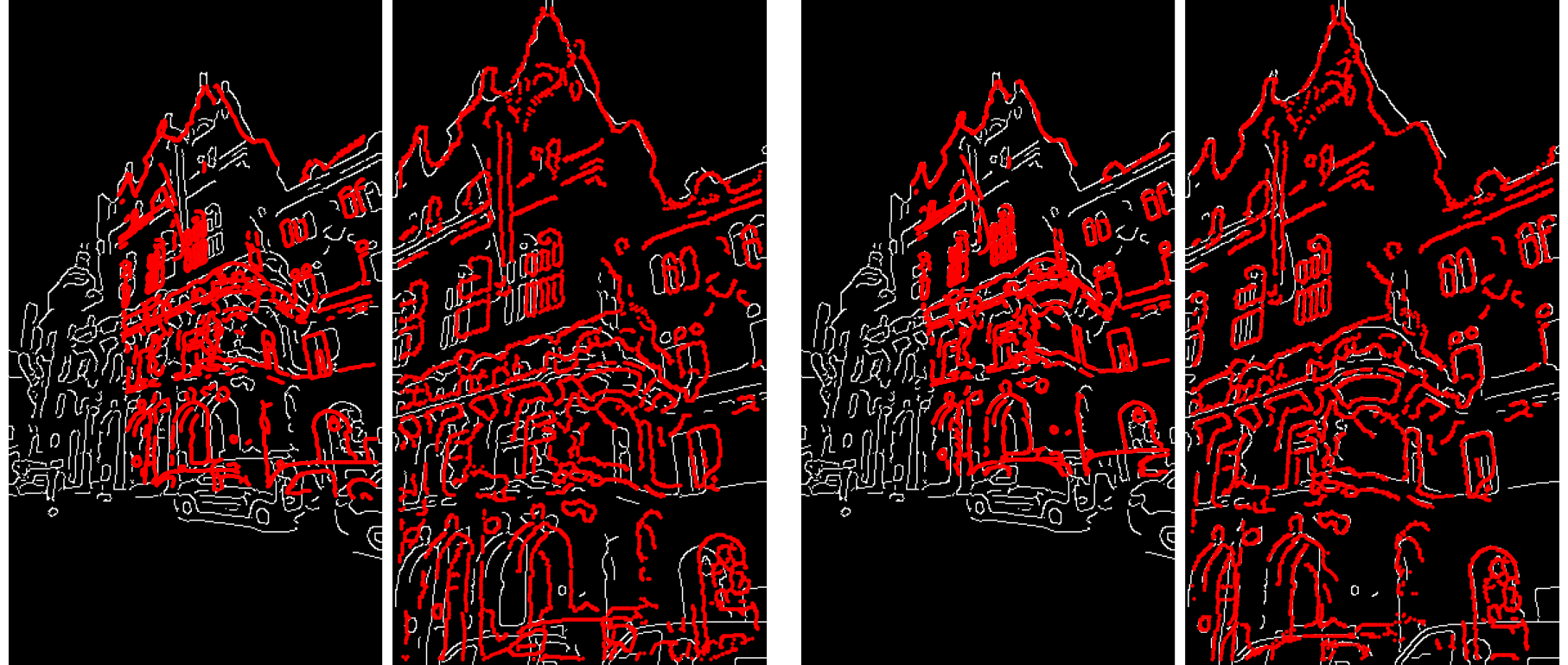}
        \caption{Bidirectional edge alignment using the initial $\mathcal{G}$ (left)
        and after refining camera parameters and depth with EPO (right).}
        \label{fig:edge_alignment}
    \end{minipage}
\end{figure}

Given an unordered set of RGB images $\mathcal{I}$, we employ a 3DFM to produce an
initial geometric estimate
\begin{equation}
    \mathcal{G} = \{(\mathbf{K}_i, \mathbf{P}_i, \mathbf{Z}_i) \mid \forall I_i \in \mathcal{I}\},
\end{equation}
as described in \cref{sec:preliminaries}. We augment $\mathcal{G}$ by computing, for
each image, the edge map $\mathbf{E}_i$ via the Canny edge
detector~\cite{canny2009computational} and the corresponding Distance Transform
Field~($\mathrm{DTF}_i$)~\cite{felzenszwalb2012distance}, in which each pixel encodes
the Euclidean distance to the nearest edge (\cref{fig:rgb_data}). Rather than relying on feature-based matching, we use edges to define a geometric
alignment measure. We exploit the dense depth maps $\mathbf{Z}_i$ provided by the
3DFM to assess pairwise pose quality from edge proximity in the image plane.

We formulate the objective as an iterative minimization problem. At each iteration $s$,
we compute the pairwise edge alignment (shown in \cref{fig:edge_alignment}) and the corresponding loss $\mathcal{L}$ over
the pairs in the viewgraph. In the first phase, this loss drives updates to the camera
intrinsics $\mathbf{K}$ and poses $\mathbf{P}$. In the subsequent joint phase,
optimization is extended to include the depth maps $\mathbf{Z}$. All updates are
computed via backpropagation using a gradient-based optimizer.

\subsubsection{Viewgraph Estimation}
\label{sec:viewgraph}
Similarly to BA, our framework requires establishing geometric correspondences across
views. However, we relax the strict requirement for explicit 2D-2D feature
correspondences, bypassing the need for feature detectors, matchers, or trackers.
Given $\mathcal{G}$, we employ an exhaustive cycle-consistent reprojection strategy
to filter out pairs with insufficient reprojected points. Specifically, we project each pixel
$\mathbf{p}$ of image $I_i$ onto $I_j$ using the projective function $\pi$:
\begin{equation}
    \mathbf{p}_{i \to j} = \pi(\mathbf{K}_i, \mathbf{K}_j, \mathbf{P}_{i},
    \mathbf{P}_{j}, \mathbf{Z}_{i}; \mathbf{p}).
\end{equation}
The resulting $\mathbf{p}_{i \to j}$ is then projected back onto $I_i$ to
obtain $\mathbf{p}_{i \to j \to i}$. An image pair is called \textit{valid} if sufficiently
many pixels satisfy a bidirectional reprojection error
$\|\mathbf{p} - \mathbf{p}_{i \to j \to i}\|_2 < \tau$. The set of \textit{valid}
pairs forms the viewgraph $\mathcal{E}$.

\subsubsection{Pose Refinement}
We employ a six-layer Multi-Layer Perceptron (MLP) with a skip connection after the
third layer, following ACE0~\cite{brachmann2024scene}. Camera rotations are represented
using the continuous 6D parameterization suggested by Zhou et al.~\cite{zhou2019continuity}. We model the
pose refinement as
\begin{equation}
    \mathbf{P}_i^s = \phi(\mathbf{P}_i^0 + \text{MLP}(\mathbf{P}_i^0)) + [\mathbf{0} \mid \delta_i],
\end{equation}
where $\mathbf{P}_i^0$ and $\mathbf{P}_i^s$ denote the camera pose matrices at
initialization and at iteration $s$, respectively. The operator $\phi$ applies
Gram-Schmidt orthonormalization to recover a valid rotation matrix from the 6D
representation. The variable $\delta_i$ is a set of learnable parameters acting as a
translation offset. We find that the MLP alone struggles to predict accurate
translation offsets, often failing to capture fine-grained adjustments; $\delta_i$ is
therefore introduced to absorb this residual error.

\subsubsection{Camera Refinement}
For images sharing a camera, we first average their focal length predictions to
establish a robust initial estimate. We then refine only the focal length via a
learnable scaling factor $\gamma_i$, such that
\begin{equation}
    f_i^s = f_i^0 \cdot (1 + \gamma_i).
\end{equation}

\subsubsection{Depth Refinement}
We refine depth \textit{pixel-wise} by learning two sets of parameters, $\alpha_i$ and
$\beta_i$, such that
\begin{equation}
    \mathbf{Z}_{i,(x,y)}^s = \mathbf{Z}_{i,(x,y)}^0 \cdot \alpha_{i,(x,y)} + \beta_{i,(x,y)}
\end{equation}
for each pixel location $(x,y)$. We empirically find that this \textit{pixel-wise} affine
parameterization yields better performance than both a single per-pixel offset and a per-image affine correction.

\subsubsection{Edge Reprojection Loss}
We define the global optimization objective as the minimization of the average edge
reprojection error across the viewgraph $\mathcal{E}$:
\begin{equation}
    \mathcal{L} = \frac{1}{|\mathcal{E}|} \sum_{(i,j) \in \mathcal{E}} \mathcal{L}_{ij},
\end{equation}
where $\mathcal{L}_{ij}$ denotes the bidirectional alignment cost for image pair
$(i, j)$:
\begin{equation}
    \mathcal{L}_{ij} = \frac{1}{|\mathbf{E}_i|} \sum_{\mathbf{p} \in \mathbf{E}_i}
    \mathcal{H}\!\left(\hat{d}_{i \to j}(\mathbf{p})\right) +
    \frac{1}{|\mathbf{E}_j|} \sum_{\mathbf{p} \in \mathbf{E}_j}
    \mathcal{H}\!\left(\hat{d}_{j \to i}(\mathbf{p})\right).
\end{equation}
Here, $\mathcal{H}$ denotes the Huber loss, and
$\hat{d}_{i \to j}(\mathbf{p}) = \min(d_{i \to j}(\mathbf{p}), \lambda)$ clamps
per-pixel distances to mitigate the influence of outliers. The per-pixel distance
$d_{i \to j}(\mathbf{p})$ is obtained by projecting each edge pixel
$\mathbf{p} \in \mathbf{E}_i$ into the target frame $I_j$ and sampling the
precomputed distance transform:
\begin{equation}
    d_{i \to j}(\mathbf{p}) = \mathrm{DTF}_j\!\left[
    \pi(\mathbf{K}_i, \mathbf{K}_j, \mathbf{P}_i, \mathbf{P}_j, \mathbf{Z}_i;
    \mathbf{p}) \right], \quad \forall\, \mathbf{p} \in \mathbf{E}_i.
    \label{eq:dtf_ij}
\end{equation}
Sampling $\mathrm{DTF}_j$ at the projected locations yields a differentiable geometric
signal that guides the optimization toward dense edge alignment.

\subsubsection{Optimization Schedule}
After each step $s$, we evaluate the change in rotation and translation at each pose
$\mathbf{P}^s_i$ with respect to the previous iteration $s-1$. The rotation change
$\Delta R_{i}^{s}$ and translation change $\Delta t_{i}^{s}$, both expressed in
degrees, are defined as:
\begin{equation}
    \Delta R_{i}^{s} = \frac{180}{\pi} \arccos\left(
    \frac{\mathrm{tr}(\mathbf{R}_i^{s} (\mathbf{R}_i^{s-1})^\top) - 1}{2} \right),
\end{equation}
\begin{equation}
    \Delta t_{i}^{s} = \frac{180}{\pi} \arccos\left(
    \frac{\mathbf{t}_i^{s}}{\lVert \mathbf{t}_i^{s} \rVert_2} \cdot
    \frac{\mathbf{t}_i^{s-1}}{\lVert \mathbf{t}_i^{s-1} \rVert_2} \right).
\end{equation}
To ensure robustness against outliers, we summarize the per-image rotation and
translation changes by their $95^{\text{th}}$ percentile over all poses,
$\Delta R^{s}_{95}$ and $\Delta t^{s}_{95}$, and define the global pose change as
$\theta^{s}_{95} = \max(\Delta R^{s}_{95}, \Delta t^{s}_{95})$.

The optimization follows a dual-phase convergence schedule designed to stabilize
camera parameters before initiating depth refinement.
\begin{enumerate}
    \item \textbf{First Convergence:} Reached when $\theta^s_{95}$ remains below
    a tolerance threshold $\epsilon_1$ for $w_1$ consecutive steps, at which point
    depth optimization is initiated.
    \item \textbf{Second Convergence:} $\theta^s_{95}$ is further monitored until
    it remains below a stricter threshold $\epsilon_2 < \epsilon_1$ for $w_2$
    consecutive steps, at which point the optimization terminates.
\end{enumerate}

\subsubsection{Implementation Details}
\label{sec:details}
We construct the viewgraph $\mathcal{E}$ by selecting image pairs for which at least $12.5\%$ of their reprojected 2D points have a bidirectional reprojection error below $\tau = 3.0$ pixels, as described in \cref{sec:viewgraph}.

The MLP pose refiner consists of 12 input units, corresponding to the flattened pose
matrix, and 12 output units representing the pose offset. The network has a constant
hidden size of 128 channels and ReLU activations. All layers use Kaiming normal
initialization~\cite{he2015delving}, except the final layer, which is initialized to
zero. All other learnable parameters are initialized to their identity values, ensuring
a neutral contribution at the first iteration that grows smoothly as optimization
progresses.


To ensure stable convergence, we apply a linear learning rate warmup from $0$ to
$3 \times 10^{-3}$ over the first 25 steps, followed by cosine annealing decay over
a maximum of $N = 2000$ iterations. Note that the output of the optimization is the
refined $\mathcal{G}$, not the MLP pose refiner itself. The framework uses \texttt{BF16} for the MLP and the Edge extraction and \texttt{FP32} precision for everything else. It is implemented in  PyTorch~\cite{paszke2019pytorch} and
Triton~\cite{tillet2019triton}. All parameters are optimized with the AdamW
optimizer~\cite{loshchilov2019decoupled} with $\beta_1 = 0.9$, $\beta_2 = 0.999$,
and weight decay $0.01$, except for the translation offset $\delta$, whose weight
decay is increased to $0.1$.

The optimization terminates based on the convergence criterion of
\cref{sec:details}, with sliding window sizes $w_1 = 25$ and $w_2 = 50$ and
tolerance levels $\epsilon_1 = 0.5$ and $\epsilon_2 = 0.1$.

\section{Experiments}
\label{sec:experiments}
\subsubsection{Testing Details}
We use VGGT~\cite{wang2025vggt} as a representative state-of-the-art 3DFM to generate
the initial pose and geometry. Nevertheless, our framework remains agnostic to the specific 3DFM, as we show in the Supplementary Material. We compare EPO against two refinement baselines introduced in VGGT:
VGGT+BA and VGGT+Ref+BA.

VGGT+BA applies BA directly to the raw VGGT output using 10 query frames and 2048
query points. As in EPO, shared camera intrinsics are averaged to provide a consistent
geometric constraint during optimization.

VGGT+Ref+BA incorporates an additional track refinement step prior to BA, consisting of an iterative update of 2D point tracks to improve correspondence accuracy. 
However, this step relies on dense feature correlation and iterative self-attention, making it computationally intensive and requiring high-end hardware with at least 40\,GB of VRAM. On the other hand, EPO runs on consumer-grade GPUs requiring only few GB of VRAM.

\subsubsection{Datasets}
\label{sec:datasets}
We evaluate across datasets representing diverse capturing scenarios:
\begin{itemize}
    \item \textbf{Indoor:} \textit{ScanNet++}~\cite{yeshwanth2023scannet++}, following
    the data split of Depth Anything 3~\cite{lin2025depth}.

    \item \textbf{Outdoor Forward-Facing:} The test set of \textit{TerraSky3D}~\cite{durso2026terrasky}, comprising large-scale scenes with
    cameras oriented toward outdoor buildings.

    \item \textbf{Object-Centric:} The scenes provided by the \textit{Mip-NeRF~360}
    dataset~\cite{barron2022mip}.
\end{itemize}

Due to the memory constraints of VGGT+BA, all scenes are capped at 150 images, equivalent to 2.5 minutes of videos at 1 fps,
uniformly sampled from their respective sorted sequences. For \textit{Mip-NeRF~360},
we retain only scenes containing at least 150 images to maximize multi-view coverage
while preserving a sufficient number of views for Novel View Synthesis
evaluation.

Methods marked with $\dagger$ are run on an NVIDIA H200 due to their higher memory
requirements. Transformer-based models benefit from the H200's increased memory
bandwidth and architectural optimizations, yielding lower inference times than on the
RTX~4090 used for all other methods. To reduce VGGT's memory footprint, we discard
intermediate layer outputs that are saved but never consumed;
this increases the maximum number of processable images from approximately 50 to 150
on the RTX~4090.

\subsection{Pose Refinement}
\label{sec:pose_eval}
For all datasets, we evaluate pose accuracy and runtime. Pose accuracy is measured
against ground-truth poses as the Area Under the recall Curve (AUC) of the maximum angular
error between the relative rotation and translation, $\max(\Delta R, \Delta T)$, in
degrees, at a threshold of $5^\circ$. Runtime is reported as the total \textit{inference} wall-clock
time excluding I/O.

\subsubsection{Indoor}
ScanNet++~\cite{yeshwanth2023scannet++} is a challenging indoor benchmark
characterized by complex geometries and varying lighting conditions. As shown in
\cref{tab:scannetpp}, the raw VGGT output is rather inaccurate on these scenes.
Both BA variants improve upon the raw results, though track refinement yields
little additional benefit. We hypothesize that this is due to the prevalence of
textureless regions and repetitive structures, which hinder reliable keypoint
detection and accurate refinement. EPO is less sensitive to these conditions, as
edges provide more informative geometric observations in feature-poor environments.
In particular, EPO improves AUC by 6 points over the best BA-based method, while
reducing execution time by ${\sim}6\times$.

\begin{table}[tp]
    \centering
    \caption{Comparison of AUC at $5^\circ$ and the total \textit{inference} wall-clock time on ScanNet++ \cite{yeshwanth2023scannet++, lin2025depth}. +BA, +Ref+BA, and +EPO are optimization methods running on top of VGGT. $\dagger$ indicates run on an NVIDIA H200. We highlight the \colorbox{best}{best} and \colorbox{second}{second best} AUC score.}
    \label{tab:scannetpp}
    \resizebox{0.6\textwidth}{!}{%
    \begin{tabular}{l cc| cc cc cc }
        \toprule
         & \multicolumn{2}{c|}{\textbf{VGGT}} & \multicolumn{2}{c}{\textbf{+BA}} & \multicolumn{2}{c}{\textbf{+Ref+BA$^\dagger$}} & \multicolumn{2}{c}{\textbf{+EPO}} \\
        \textbf{Scene} & AUC$\uparrow$ & Time$\downarrow$ & AUC$\uparrow$ & Time$\downarrow$ & AUC$\uparrow$ & Time$\downarrow$ & AUC$\uparrow$ & Time$\downarrow$ \\
        \hline
        09C1414F1B & 40.1 & 40.3 & 66.4 & 121.4 & \cellcolor{second}66.8 & 335.6 & \cellcolor{best}{72.3} & 54.3 \\
        1Ada7A0617 & 48.5 & 36.8 & 64.0 & 129.3 & \cellcolor{second}64.8 & 304.4 & \cellcolor{best}{66.5} & 49.4 \\
        21D970D8De & 66.1 & 37.1 & \cellcolor{second}76.8 & 118.6 & 73.5 & 251.7 & \cellcolor{best}{83.5} & 53.2 \\
        286B55A2Bf & 48.0 & 37.2 & 72.4 & 394.8 & \cellcolor{second}72.9 & 584.3 & \cellcolor{best}{76.7} & 54.3 \\
        38D58A7A31 & 61.2 & 36.7 & 75.1 & \phantom{0}96.7 & \cellcolor{second}76.8 & 282.0 & \cellcolor{best}{81.4} & 50.2 \\
        3E8Bba0176 & 73.0 & 36.9 & \cellcolor{second}76.7 & 117.7 & 75.5 & 267.0 & \cellcolor{best}{83.7} & 50.3 \\
        40Aec5Fffa & 45.8 & 36.7 & 50.7 & 101.0 & \cellcolor{second}51.4 & 229.8 & \cellcolor{best}{72.4} & 50.1 \\
        578511C8A9 & 50.9 & 36.8 & 64.6 & 298.4 & \cellcolor{second}65.3 & 497.3 & \cellcolor{best}{68.8} & 55.5 \\
        5F99900F09 & 44.9 & 36.7 & \cellcolor{second}71.8 & 107.6 & 71.1 & 230.5 & \cellcolor{best}{76.4} & 49.4 \\
        7831862F02 & 70.1 & 37.1 & 80.0 & 193.5 & \cellcolor{second}80.8 & 218.6 & \cellcolor{best}{87.6} & 48.3 \\
        7Bc286C1B6 & 50.3 & 36.7 & 62.2 & 228.9 & \cellcolor{second}62.9 & 258.0 & \cellcolor{best}{71.0} & 48.7 \\
        9071E139D9 & 64.9 & 36.9 & 73.3 & 108.4 & \cellcolor{best}{75.9} & 179.8 & \cellcolor{second}{74.1} & 50.8 \\
        Acd95847C5 & 69.3 & 37.3 & 70.8 & 175.7 & \cellcolor{second}{74.5} & 352.4 & \cellcolor{best}{79.8} & 50.4 \\
        Bcd2436Daf & 53.8 & 38.2 & 73.2 & 209.0 & \cellcolor{second}{78.4} & 267.6 & \cellcolor{best}{82.7} & 61.7 \\
        Bde1E479Ad & 56.8 & 37.5 & 70.0 & 248.1 & \cellcolor{second}{73.5} & 340.8 & \cellcolor{best}{80.8} & 46.5 \\
        C4C04E6D6C & 50.2 & 36.4 & 68.2 & 181.0 & \cellcolor{second}{69.9} & 405.2 & \cellcolor{best}{78.9} & 59.1 \\
        C5439F4607 & 48.5 & 36.9 & 68.2 & 218.6 & \cellcolor{second}{69.1} & 250.2 & \cellcolor{best}{73.8} & 53.3 \\
        Cc5237Fd77 & 46.9 & 36.9 & 67.1 & 157.1 & \cellcolor{second}{67.8} & 348.2 & \cellcolor{best}{72.9} & 50.7 \\
        F3D64C30F8 & 57.2 & 36.1 & \cellcolor{second}74.0 & 118.5 & {69.0} & 248.4 & \cellcolor{best}{77.6} & 54.1 \\
        Fb5A96B1A2 & 66.0 & 37.1 & 75.2 & 108.1 & \cellcolor{second}{77.1} & 217.6 & \cellcolor{best}{81.4} & 51.9 \\
        \midrule
        \textbf{Mean} & 55.6 & 37.1 & 70.0 & 171.6 & \cellcolor{second}{70.9} & 303.5 & \cellcolor{best}{77.1} & 52.1 \\
        \bottomrule
    \end{tabular}}
\end{table}

\subsubsection{Outdoor}
TerraSky3D~\cite{durso2026terrasky} is a recent dataset of popular European landmarks,
whose test set comprises nine large-scale outdoor scenes captured from a ground
perspective with various iPhone cameras. 

As reported in \cref{tab:terrasky}, VGGT accuracy is similar to the indoor case, yet
both BA-based approaches are considerably more effective here. We attribute this to the
greater texture prevalence in outdoor scenes, which enables more reliable track
refinement and improves the initial camera trajectories. Nevertheless, EPO improves
AUC by ${\sim}4$ points over the best BA-based method, adding only an average of ${\sim}14$ seconds on top of VGGT runtime.

\begin{table}[tp]
    \centering
    \caption{Comparison of AUC at $5^\circ$ and the total \textit{inference} wall-clock time on TerraSky3D \cite{durso2026terrasky}. +BA, +Ref+BA, and +EPO are optimization methods running on top of VGGT. $\dagger$ indicates run on an NVIDIA H200. We highlight the \colorbox{best}{best} and \colorbox{second}{second best} AUC score.}
    \label{tab:terrasky}
    \resizebox{0.67\textwidth}{!}{%
    \begin{tabular}{l cc | cc cc cc }
        \toprule
         & \multicolumn{2}{c|}{\textbf{VGGT}} & \multicolumn{2}{c}{\textbf{+BA}} & \multicolumn{2}{c}{\textbf{+Ref+BA$^\dagger$}} & \multicolumn{2}{c}{\textbf{+EPO}} \\
        \textbf{Scene} & AUC$\uparrow$ & Time$\downarrow$ & AUC$\uparrow$ & Time$\downarrow$ & AUC$\uparrow$ & Time$\downarrow$ & AUC$\uparrow$ & Time$\downarrow$ \\
        \hline
        Graz Church & 72.1 & 37.9 & 81.6 & 929.7 & \cellcolor{second}87.4 & 1051.4 & \cellcolor{best}{90.5} & 51.0 \\
        Graz Townhall & 18.8 & 42.4 & 65.6 & 107.1 & \cellcolor{second}65.9 & 229.0 & \cellcolor{best}{66.7} & 63.9 \\
        Graz University & 35.5 & 37.9 & 41.5 & 256.1 & \cellcolor{second}41.8 & 374.6 & \cellcolor{best}{50.5} & 56.6 \\
        Munich Frauenkirche & 68.2 & 30.0 & 83.5 & 107.0 & \cellcolor{second}87.7 & 206.1 & \cellcolor{best}{89.7} & 46.0 \\
        Munich Marienplatz & 55.3 & 40.1 & \cellcolor{second}70.5 & 109.2 & \cellcolor{best}{78.6} & 257.5 & 68.1 & 46.2 \\
        Munich Theatinerkirche & 67.0 & 28.0 & 71.6 & 151.6 & \cellcolor{second}77.7 & 341.7 & \cellcolor{best}{89.4} & 42.6 \\
        Salzburg Andr\"akirche & 70.6 & 26.1 & 75.9 & 112.0 & \cellcolor{second}76.1 & 203.1 & \cellcolor{best}{85.3} & 29.6 \\
        Salzburg Rechte Altstadt & 68.8 & 9.5 & 86.4 & 42.1 & \cellcolor{best}{93.5} & 91.1 & \cellcolor{second}90.7 & 20.8 \\
        Vienna State Opera & 54.9 & 26.7 & 63.2 & 219.7 & \cellcolor{second}70.9 & 284.5 & \cellcolor{best}{82.0} & 44.9 \\
        \midrule
        \textbf{Mean} & 56.8 & 31.0 & 71.1 & 226.1 & \cellcolor{second}75.5 & 337.7 & \cellcolor{best}{79.2} & 44.6 \\
        \bottomrule
    \end{tabular}}
\end{table}

\subsubsection{Object-Centric}
Mip-NeRF~360~\cite{barron2022mip} is an object-centric dataset primarily used for
NVS evaluation. As with the previous datasets, geometric optimization significantly
improves upon the raw VGGT output. Also in this setting, VGGT+EPO achieves higher
geometric accuracy than BA-based approaches, while delivering a $4\times$ speedup
on a consumer-grade GPU at a fraction of the cost, as shown in
\cref{tab:mipnerf360}. 

\begin{table}[tp]
    \centering
    \caption{Comparison of AUC at $5^\circ$ and the total \textit{inference} wall-clock time on Mip-NeRF 360 \cite{barron2022mip}. +BA, +Ref+BA, and +EPO are optimization methods running on top of VGGT. $\dagger$ indicates run on an NVIDIA H200. We highlight the \colorbox{best}{best} and \colorbox{second}{second best} AUC score.}
    \label{tab:mipnerf360}
    \resizebox{0.6\textwidth}{!}{%
    \begin{tabular}{l cc| cc cc cc }
        \toprule
         & \multicolumn{2}{c|}{\textbf{VGGT}} & \multicolumn{2}{c}{\textbf{+BA}} & \multicolumn{2}{c}{\textbf{+Ref+BA$^\dagger$}} & \multicolumn{2}{c}{\textbf{+EPO}} \\
        \textbf{Scene} & AUC$\uparrow$ & Time$\downarrow$ & AUC$\uparrow$ & Time$\downarrow$ & AUC$\uparrow$ & Time$\downarrow$ & AUC$\uparrow$ & Time$\downarrow$ \\
        \hline
        Bicycle & 77.8 & 44.0 & 83.4 & 88.6 & \cellcolor{second}86.5 & 179.8 & \cellcolor{best}{90.0} & 54.6 \\
        Bonsai & 69.7 & 39.8 & 87.7 & 103.1 & \cellcolor{second}87.8 & 215.6 & \cellcolor{best}{90.9} & 46.2 \\
        Counter & 83.5 & 40.8 & \cellcolor{second}91.5 & 128.2 & \cellcolor{best}{94.7} & 249.3 & \cellcolor{best}{94.7} & 45.3 \\
        Flowers & 34.3 & 43.0 & 70.2 & 91.6 & \cellcolor{second}70.7 & 177.2 & \cellcolor{best}{78.7} & 59.7 \\
        Garden & 84.8 & 43.6 & 92.2 & 85.7 & \cellcolor{best}{95.4} & 143.5 & \cellcolor{second}92.7 & 49.3 \\
        Kitchen & 87.8 & 41.0 & 87.2 & 117.1 & \cellcolor{second}88.7 & 224.0 & \cellcolor{best}{94.8} & 45.1 \\
        Room & 68.6 & 41.1 & \cellcolor{second}88.5 & 113.0 & \cellcolor{best}{91.8} & 281.9 & \cellcolor{best}{91.8} & 47.3 \\
        \midrule
        \textbf{Mean} & 72.2 & 41.9 & 85.5 & 103.9 & \cellcolor{second}87.8 & 210.2 & \cellcolor{best}{90.5} & 49.6 \\
        \bottomrule
    \end{tabular}}
\end{table}

\subsection{Reprojection Error}

\cref{tab:model_performance} reports the average median reprojection error (RE) for
the sparse point clouds and the average 2D observation count (Obs.) across all
datasets. We define an observation as any 2D image feature contributing to the
optimization, either a detected keypoint (for BA-based methods) or an edge pixel
(for EPO). For BA-based variants, RE is the $L_2$ distance between the projected
3D point and its associated 2D keypoint.

EPO does not rely on explicit 3D-2D correspondences; instead, we assess geometric
consistency via bidirectional 2D-2D reprojection. Following \cref{eq:dtf_ij}, we
reproject edges across views and sample the $\mathrm{DTF}$ at the target locations,
using these distances as a proxy for RE. Given the high density of such observations,
we report the median RE averaged first per scene and then across each dataset.

EPO achieves a comparable mean RE of $2.0$\,px to Ref+BA, while producing $1.7\times$ more 2D observations than the BA baselines.

\begin{table}[h]
    \centering
    \caption{Average median RE (in pixels) and 2D observations (in thousands) per
    scene. RE is the raw reprojection error (no robustification). Best results are in \colorbox{best}{green}.}
    \label{tab:model_performance}
    \small
    \resizebox{0.7\linewidth}{!}{
    \begin{tabular}{l cc cc cc cc}
        \toprule
        & \multicolumn{2}{c}{\textbf{TerraSky3D}}
        & \multicolumn{2}{c}{\textbf{ScanNet++}}
        & \multicolumn{2}{c}{\textbf{Mip-NeRF~360}}
        & \multicolumn{2}{c}{\textbf{Mean}}\\
        \cmidrule(lr){2-3} \cmidrule(lr){4-5} \cmidrule(lr){6-7} \cmidrule(lr){8-9}
        VGGT & RE$\downarrow$ & Obs.$\uparrow$ & RE$\downarrow$ & Obs.$\uparrow$
             & RE$\downarrow$ & Obs.$\uparrow$ & RE$\downarrow$ & Obs.$\uparrow$ \\
        \cmidrule(r){1-7} \cmidrule(l){8-9}
        \rotatebox[origin=c]{180}{$\Lsh$} +BA
            & 3.7 & 102 & 3.4 & 24 & 1.6 & 107 & 2.9 & \phantom{0}78\\
        \rotatebox[origin=c]{180}{$\Lsh$} +Ref+BA
            & 2.6 & 102 & 2.2 & 25 & \cellcolor{best}{1.3} & 109 & \cellcolor{best}{2.0} & \phantom{0}79\\
        \rotatebox[origin=c]{180}{$\Lsh$} +EPO
            & \cellcolor{best}{1.2} & \cellcolor{best}{150} & \cellcolor{best}{1.9} & \cellcolor{best}{60}
            & 3.0 & \cellcolor{best}{199} & \cellcolor{best}{2.0} & \cellcolor{best}{136}\\
        \bottomrule
    \end{tabular}
    }
\end{table}

\subsection{Novel View Synthesis}
\begin{table}[t]
    \centering
    \caption{Novel View Synthesis evaluation on the Mip-NeRF 360 \cite{barron2022mip} dataset. All models were rendered using 3DGS \cite{kerbl20233d} trained for 30,000 steps. Performance is quantified in terms of PSNR, SSIM, and LPIPS using standard test-set sampling. In \colorbox{best}{green} the best mean scores for each metric.}
    \label{tab:nvs}
    \resizebox{\linewidth}{!}{
         \begin{tabular}{l ccc | ccc ccc ccc ccc}
            \toprule
              & \multicolumn{3}{c}{\textbf{GT}} & \multicolumn{3}{c}{\textbf{VGGT}} & \multicolumn{3}{c}{\textbf{VGGT+BA}} & \multicolumn{3}{c}{\textbf{VGGT+Ref+BA}} & \multicolumn{3}{c}{\textbf{VGGT+EPO}} \\
            \cmidrule(lr){2-4} \cmidrule(lr){5-7} \cmidrule(lr){8-10} \cmidrule(lr){11-13} \cmidrule{14-16}

            \textbf{Scene} & \textbf{PSNR$\uparrow$} & \textbf{SSIM$\uparrow$} &\textbf{LPIPS$\downarrow$} & \textbf{PSNR$\uparrow$} & \textbf{SSIM$\uparrow$} &\textbf{LPIPS$\downarrow$} & \textbf{PSNR$\uparrow$} & \textbf{SSIM$\uparrow$} &\textbf{LPIPS$\downarrow$} & \textbf{PSNR$\uparrow$} & \textbf{SSIM$\uparrow$} &\textbf{LPIPS$\downarrow$} & \textbf{PSNR$\uparrow$} & \textbf{SSIM$\uparrow$} &\textbf{LPIPS$\downarrow$} \\

            \hline
            Bicycle & 21.77 & 0.685 & 0.271 & 16.77 & 0.357 & 0.524 & 18.74 & 0.453 & 0.444 & 19.43 & 0.488 & 0.423 & 19.25 & 0.504 & 0.399 \\
            Bonsai & 30.14 & 0.931 & 0.207 & 21.96 & 0.682 & 0.414 & 25.36 & 0.804 & 0.310 & 25.50 & 0.816 & 0.303 & 27.60 &  0.874 & 0.253 \\
            Counter & 27.98 & 0.896 & 0.195 & 22.98 & 0.719 & 0.334 & 25.49 & 0.814 & 0.278 & 26.33 & 0.847 & 0.249 &  25.87 & 0.834 & 0.240 \\
            Flowers & 20.69 & 0.624 & 0.319 & 14.86 & 0.269 & 0.612 & 17.41 & 0.375 & 0.467 & 17.32 & 0.380 & 0.467 &  19.09 & 0.505 & 0.386\\
            Garden & 27.03 & 0.848 & 0.128 & 20.80 & 0.467 & 0.387 & 22.86 & 0.669 & 0.270 & 23.73 & 0.733 & 0.222 & 23.27 & 0.685 & 0.234\\
            Kitchen & 28.83 & 0.927 & 0.116 & 20.69 & 0.595 & 0.333 & 20.61 & 0.652 & 0.333 & 20.73 & 0.677 & 0.320 &  23.37 & 0.776 & 0.201\\
            Room & 30.39 & 0.903 & 0.229 & 25.86 & 0.797 & 0.349 & 28.40 & 0.859 & 0.275 & 28.40 & 0.873 & 0.258 &  29.06 & 0.867 & 0.274\\
            \hline
            \textbf{Mean} & 26.69 & 0.831 & 0.209 & 20.56 & 0.555 & 0.422 & 22.70 & 0.661 & 0.339 & 23.06 & 0.688 & 0.320 & \cellcolor{best}{23.93} & \cellcolor{best}{0.721} & \cellcolor{best}{0.284} \\
            \bottomrule
        \end{tabular}
        }
       
\end{table}

Recently, due to a scarcity of real-world datasets with ground-truth poses, several
works~\cite{brachmann2024scene, wang2025vggt, zhong2025instantsfm} have proposed
evaluating SfM reconstruction quality via downstream tasks, the most common being
Novel View Synthesis (NVS)~\cite{waechter2017virtual}. This involves rendering
frameworks such as Neural Radiance Fields (NeRF)~\cite{mildenhall2021nerf} or 3D
Gaussian Splatting (3DGS)~\cite{kerbl20233d} to reconstruct a scene from the camera
poses estimated by the SfM pipeline. The quality of the synthesized views then serves
as a proxy for the accuracy and consistency of the underlying geometric reconstruction:
high-fidelity rendering indicates that the estimated camera trajectories and scene
geometry satisfy the strict multi-view constraints required for photorealistic
synthesis.

In our experiments, we train a 3DGS~\cite{kerbl20233d} model for 30,000 steps on
each SfM reconstruction, following the data split of~\cite{kerbl20233d}. Note that
COLMAP pseudo-GT values are lower than those reported in~\cite{kerbl20233d} because
the scenes are downsampled as described in \cref{sec:datasets}, slightly reducing
visual coverage.

As shown in \cref{tab:nvs}, VGGT+EPO outperforms VGGT+Ref+BA on all metrics,
consistent with the geometric accuracy reported in \cref{tab:mipnerf360}.
\cref{fig:nvs_render} further confirms this trend qualitatively: images rendered from
VGGT+EPO reconstructions exhibit superior visual fidelity, with sharp structural
details and high-frequency background textures well preserved, whereas renders from
both BA-based methods show visible degradation.

\begin{table}[t]
    \centering
    \small
    \begin{minipage}[t]{0.45\textwidth}
        \centering
        \caption{Ablation study on Mip-NeRF 360 evaluating LPIPS and AUC at $5^\circ$ threshold. Time (in seconds) refers to the optimization only duration.}
        \label{tab:ablation} 
        \resizebox{\linewidth}{!}{
            \begin{tabular}{l r | c c c}
                \toprule
                & Steps & LPIPS$\downarrow$ & AUC$\uparrow$ & Time$\downarrow$ \\
                \hline
                VGGT (baseline) & 0 & 0.422 & 72.2 & \phantom{0}0 \\
                \hline
                Free Variables  & 100 & 0.343 & 77.0 & \phantom{0}4  \\            
                \rotatebox[origin=c]{180}{$\Lsh$} w/ Pose MLP &                     100 & 0.339 & 79.0 & \phantom{0}4  \\
                \rotatebox[origin=c]{180}{$\Lsh$} w/ Parametric $K$ and $Z$ &       100 & 0.323 & 80.2 & \phantom{0}4  \\
                \rotatebox[origin=c]{180}{$\Lsh$} w/ Translation Offset $\delta$ &  100 & 0.308 & 82.0 & \phantom{0}4  \\
                \hline
                \rotatebox[origin=c]{180}{$\Lsh$} w/ $10\times$ Iterations & 1000 & 0.269 & 91.0 & 14  \\
                \rotatebox[origin=c]{180}{$\Lsh$} w/ $20\times$ Iterations & 2000 & 0.266 & 92.1 & 25  \\
                \rotatebox[origin=c]{180}{$\Lsh$} w/ Stopping Criterion & \textit{avg.} 387 & 0.284 & 90.5 & \phantom{0}7  \\ 
                \bottomrule
            \end{tabular}
        }
    \end{minipage}
    \hfill 
    \begin{minipage}[t]{0.53\textwidth}
        \centering
        \caption{Ablation study on stopping criteria. Max refers to the maximum AUC at $5^\circ$ threshold reached over 2,000 optimization steps. Time (in seconds) includes optimization runtime only.}
        \label{tab:early_stop} 
        \resizebox{\linewidth}{!}{
            \begin{tabular}{l c|cc cc cc}
                \toprule
                & \multicolumn{1}{c}{Max} & \multicolumn{2}{c}{Fixed} & \multicolumn{2}{c}{Loss-based} & \multicolumn{2}{c}{Pose-based} \\
                \cmidrule(lr){2-2}  \cmidrule(lr){3-4}  \cmidrule(lr){5-6} \cmidrule(lr){7-8} 
                & AUC$\uparrow$ & AUC$\uparrow$ & Time$\downarrow$ &  AUC$\uparrow$ & Time$\downarrow$ &  AUC$\uparrow$ & Time$\downarrow$ \\
                \hline
                TerraSky3D    & 80.1 & 80.0 & 24 & 76.0 & \phantom{0}7 & 79.2 & 14 \\
                ScanNet++     & 78.3 & 77.1 & 25 & 76.9 & 16 & 77.1 & 15 \\
                Mip-NeRF 360  & 92.3 & 91.1 & 25 & 89.5 & \phantom{0}9 & 90.5 & \phantom{0}7 \\
                \hline 
                Mean          & 83.6 & 82.7 & 25 & 80.8 & 11 & 82.2 & 12 \\
                \bottomrule
            \end{tabular}
        }
    \end{minipage}
\end{table}

\subsection{Ablation Study}

\subsubsection{Design and Runtime}
To motivate our design choices, we conduct a two-stage ablation study evaluating pose
accuracy (AUC at $5^\circ$), perceptual quality of rendered images (LPIPS~\cite{zhang2018unreasonable}),
and inference time. 

The initial formulation, which optimizes the camera parameters in $\mathcal{G}$ as
free variables, frequently becomes unstable beyond 100 iterations. We therefore first
isolate the contribution of individual components within this low-iteration regime,
then evaluate the full system at higher step counts using the more stable
configurations. All configurations follow a two-stage schedule: the first stage
stabilizes camera parameters, after which optimization continues for a fixed number
of steps including depth refinement. The only exception is the last row, which uses
our proposed adaptive stopping criterion.

As detailed in \cref{tab:ablation}, our baseline optimizes the elements of
$\mathcal{G}$ (encoded as \textit{q}, \textit{t}, $\log\frac{f}{W}$,
$\frac{1}{Z}$) as free variables. We then incrementally add the MLP pose refiner,
the reparameterizations of $\mathbf{K}$ and $\mathbf{Z}$ described in
\cref{sec:method}, and finally the translation offset $\delta$. Each addition yields
significant gains in pose accuracy and rendering quality at negligible computational
cost.

The lower portion of \cref{tab:ablation} reports the effect of extended optimization.
While increasing the iteration count brings incremental gains, it introduces
substantial computational overhead. Our adaptive stopping criterion effectively
addresses this trade-off, reducing the optimization runtime by approximately 50\% and 70\%
relative to the 1000- and 2000-iteration baselines, respectively, while maintaining
comparable geometric accuracy and perceptual fidelity. Since complex scenes often
require more iterations to converge, this criterion is essential for ensuring
efficiency across diverse environments without sacrificing reconstruction quality.

\begin{figure*}[t]
     \centering
     \begin{subfigure}[b]{0.325\textwidth}
         \centering
         \includegraphics[width=\textwidth]{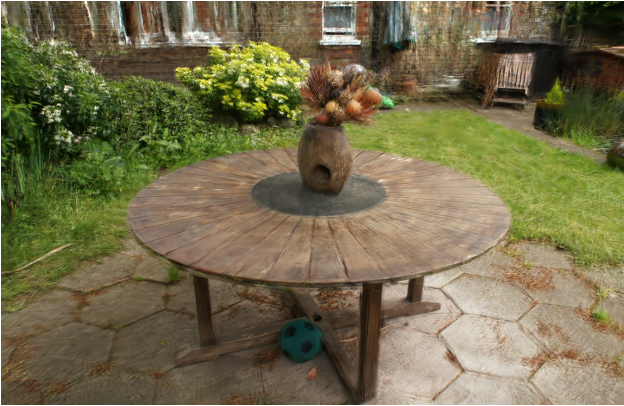}
         \caption{VGGT}
         \label{fig:image1}
     \end{subfigure}
     \hfill 
     \begin{subfigure}[b]{0.325\textwidth}
         \centering
         \includegraphics[width=\textwidth]{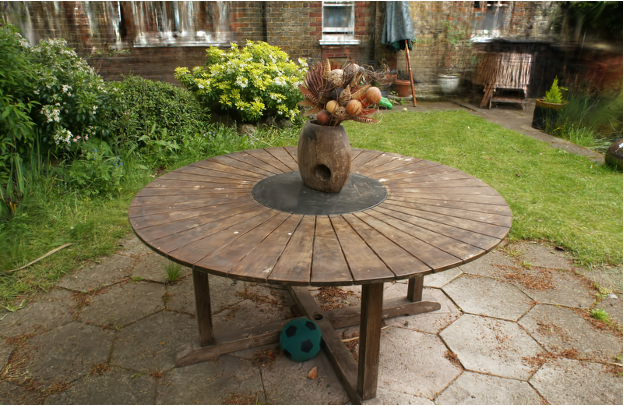}
         \caption{VGGT+Ref+BA}
         \label{fig:image2}
     \end{subfigure}
     \hfill 
     \begin{subfigure}[b]{0.325\textwidth}
         \centering
         \includegraphics[width=\textwidth]{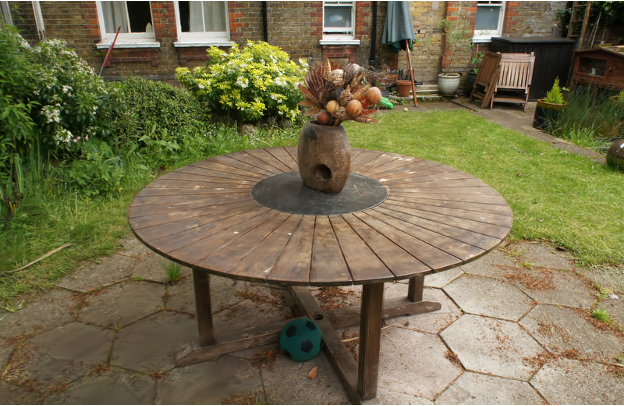}
         \caption{VGGT+EPO (Ours)}
         \label{fig:image3}
     \end{subfigure}
     
    \caption{\textbf{Qualitative Novel View Synthesis Results.} Example renderings from the \textit{Garden} scene of the Mip-NeRF 360 dataset \cite{barron2022mip}. All 3DGS models were initialized using SfM reconstructions produced by their respective methods. (a) Base VGGT output. (b) Refined reconstruction utilizing track refinement and BA (VGGT+Ref+BA). (c) Our proposed edge-based optimization (VGGT+EPO). Notably, our approach preserves sharper structural details in the background compared to the BA-based refinement.}
    \label{fig:nvs_render}
\end{figure*}

\subsubsection{Stopping Criteria}
\label{sec:stopping_criteria}
\cref{tab:early_stop} reports AUC at $5^\circ$ and runtime of EPO across datasets under various stopping criteria, with a maximum budget of 2000 iterations. The leftmost column presents the maximum AUC achieved by running EPO for the full 2000 iterations and sampling the output every 50 steps; it therefore represents an empirical upper bound on achievable accuracy.

The remaining three columns correspond to EPO configurations using a \textit{fixed} iteration count, a \textit{loss-based} stopping criterion, and our proposed \textit{pose-based} stopping criterion. As described in \cref{sec:method}, the optimization follows two stages: the first optimizes camera intrinsics and extrinsics only, while the second adds depth. All criteria share the same first stage; the distinction lies in how the second stage is terminated. The \textit{fixed} approach exhausts the full iteration budget after the first stage. The \textit{loss-based} criterion monitors relative changes in the loss, stopping when it remains below a threshold $\epsilon_2 = 5\times10^{-4}$ for $w_2 = 100$ consecutive steps. Our \textit{pose-based} criterion instead monitors pose convergence as described in \cref{sec:method}.

As shown in \cref{tab:early_stop}, our pose-based criterion maintains geometric accuracy comparable to the full budget (82.2 vs.\ 83.6 AUC). It matches the accuracy of the \textit{fixed} configuration (82.2 vs.\ 82.7 AUC) while reducing runtime by $2\times$, and improves over the \textit{loss-based} one (82.2 vs.\ 80.8 AUC) at comparable runtime.

\subsubsection{Time Breakdown}
\begin{figure}
    \centering 
    \includegraphics[width=0.8\linewidth]{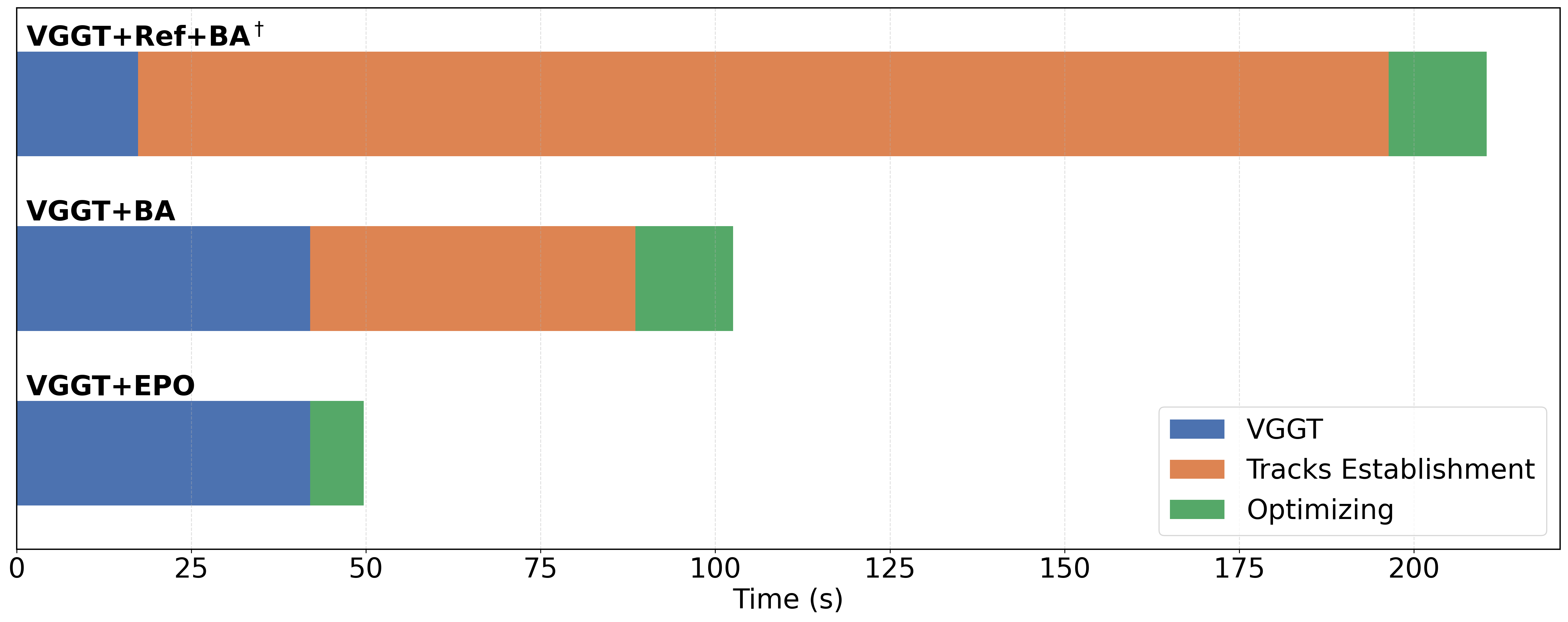}
    \caption{Absolute and relative inference time breakdown for the optimization methods
    evaluated in \cref{tab:mipnerf360}. $\dagger$ indicates runs on an NVIDIA H200.}
    \label{fig:timings}
\end{figure}
\cref{fig:timings} breaks down the per-stage runtime of VGGT+BA, VGGT+Ref+BA,
and VGGT+EPO on Mip-NeRF~360. For BA-based methods, track computation constitutes the most time-consuming stage. By contrast, EPO's trackless design, entirely eliminates this stage, allowing it to
outpace both BA-based baselines despite relying on a first-order optimizer rather than the second-order Levenberg--Marquardt solver used by BA.

The efficiency advantage of our method is further highlighted by a cross-hardware
comparison. Running VGGT on an NVIDIA H200 yields a nearly $3\times$ speedup, yet VGGT+Ref+BA$^\dagger$ remains approximately $4\times$ slower than EPO running on an
NVIDIA RTX 4090 due to the track establisment process.
\section{Discussion and Limitations}
\label{sec:limitations}
\subsection{Boosting \textit{Different} 3D Foundation Models}

We conducted the same experiments described in \cref{sec:pose_eval} using various 3DFMs reconstructions as a starting point to demonstrate our method's generalization across different models. Aside from the RGB images used for edge extraction, EPO requires only camera parameters (focal length and pose) and depth maps, which are all provided by the considered models. 

Furthermore, we disabled any masking of output depth maps to align with VGGT raw output format. This also ensures that sufficient depth information is available on detected edges.

In \cref{tab:SM_pose_eval}, we report the average AUC at $3^\circ$ and $5^\circ$ for the TerraSky3D, ScanNet++, and Mip-NeRF 360 datasets. We supplement the VGGT \cite{wang2025vggt} results with evaluations for MapAnything \cite{keetha2025mapanything} and Pi3 \cite{wang2025pi} leveraging \cite{keetha2025mapanything}'s modular software framework. As shown by these results, EPO consistently improves geometric accuracy across all tested 3DFMs by a significant margin. 

\begin{table}[htbp]
    \centering
    
    \caption{Pose evaluation in terms of AUC at $3^\circ$ and $5^\circ$. For each model, the first row measures model raw output performance, while the second row the output optimized with EPO. The best results are highlighted in \colorbox{best}{green}.}
    \label{tab:SM_pose_eval}
    \setlength{\tabcolsep}{4pt}
    \resizebox{0.8\linewidth}{!}{
        \begin{tabular}{l cc | cc | cc}
            \toprule
             & \multicolumn{2}{c|}{\textbf{TerraSky3D}} & \multicolumn{2}{c|}{\textbf{ScanNet++}} & \multicolumn{2}{c}{\textbf{Mip-NeRF 360}} \\
            & AUC@3  & AUC@5  & AUC@3  & AUC@5  & AUC@3  & AUC@5  \\
            \hline
            VGGT \cite{wang2025vggt}                 & 39.7 & 56.8 & 38.9 & 55.6 & 59.5 & 72.2 \\
            \rotatebox[origin=c]{180}{$\Lsh$} w/ EPO & \cellcolor{best}{69.4} & \cellcolor{best}{79.2} & \cellcolor{best}{64.6} & \cellcolor{best}{77.1} & \cellcolor{best}{84.3} & \cellcolor{best}{90.5} \\
            \hline
            MapAnything \cite{keetha2025mapanything} & 35.0 & 53.0 & 17.9 & 37.3 & 23.9 & 41.4 \\
            \rotatebox[origin=c]{180}{$\Lsh$} w/ EPO & \cellcolor{best}{68.1} & \cellcolor{best}{78.3} & \cellcolor{best}{41.3} & \cellcolor{best}{60.4} & \cellcolor{best}{64.9} & \cellcolor{best}{74.5} \\
             \hline
            Pi3 \cite{wang2025pi} & 48.3 & 65.1 & 55.0 & 70.0 & 60.4 & 71.8 \\
            \rotatebox[origin=c]{180}{$\Lsh$} w/ EPO & \cellcolor{best}{72.0} & \cellcolor{best}{81.2} & \cellcolor{best}{68.9} & \cellcolor{best}{80.3} & \cellcolor{best}{76.8} & \cellcolor{best}{85.9} \\

            \bottomrule
        \end{tabular}
    }
    
\end{table}

\subsubsection{Implicit Track Regularization and Limitations}
Our objective couples the edge reprojection loss across the entire viewgraph
$\mathcal{E}$, so that each focal length, camera pose, and depth value is constrained
by all pairs in which it participates. This implicitly regularizes the optimization
toward multi-view consistency and prevents localized drift without explicit,
memory-intensive tracking modules.

However, the effectiveness of this implicit regularization depends on the underlying
scene geometry and viewgraph topology. We identify two primary limitations:

\begin{description}
    \item[\textbf{Texture Sensitivity:}] Reliance on edge maps makes the optimization
    vulnerable in scenes dominated by high-frequency, non-structural textures
    (e.g., dense foliage or reflections), where edge detectors tend to extract
    inconsistent noise rather than stable structural primitives, degrading the gradient
    signal and hindering convergence.

    \item[\textbf{Viewgraph Topology:}] Viewgraphs with low node connectivity may
    provide insufficient geometric constraints to resolve ambiguities during
    optimization. This limitation depends on connection density rather than the
    absolute number of images in the scene.
\end{description}

\section{Conclusion}
We presented EPO, a trackless optimization framework and a lightweight alternative to
traditional Bundle Adjustment. By leveraging edge map alignment, EPO guides the
geometric refinement of 3D reconstructions without requiring explicit feature tracks,
bypassing the computationally expensive and memory-intensive tracking modules of
state-of-the-art 3DFMs that often demand server-grade hardware and dominate total
inference time. 

Our extensive evaluation across TerraSky3D, ScanNet++, and Mip-NeRF~360 demonstrates
that EPO achieves competitive or superior geometric accuracy compared to BA-based
pipelines while significantly reducing their computational overhead. We further
validated the quality of the refined poses through Novel View Synthesis, where EPO
consistently preserved sharper structural details. Ultimately, EPO demonstrates that
first-order gradient-based optimization, coupled with dense edge priors, offers a
fast, accessible, and trackless path toward high-fidelity 3D geometric refinement on
consumer-grade hardware.

\subsubsection{Acknowledgements} This work has been supported by the FFG under Contract No. 881844 within the project ``Pro²Future''. We also thank Felix Windisch for all the insightful discussions on Neural Rendering.
\clearpage

\title{Supplementary Material for:\\ Boosting 3D Foundation Models with \\ Edge-based Pose Optimization} 

\titlerunning{Supplementary Material: Boosting 3D Foundation Models...}

\author{
    Mattia D'Urso\inst{1}\orcidlink{0009-0003-3672-6559} \quad 
    Christian Sormann\inst{2}\orcidlink{0000-0002-6824-4007} \quad 
    Mattia Rossi\inst{2}\orcidlink{0000-0001-5158-2395} \quad  \\
    Friedrich Fraundorfer\inst{1}\orcidlink{0000-0002-5805-8892}
}

\authorrunning{M.~D'Urso et al.}

\institute{$^{1}$Graz University of Technology \quad $^{2}$Sony Europe}
\maketitle

\setcounter{page}{1}
\setcounter{figure}{0} 
\renewcommand{\thefigure}{S\arabic{figure}} 
\setcounter{table}{0}
\renewcommand{\thetable}{S\arabic{table}}

\renewcommand{\thesection}{A\arabic{section}}
\section{Pseudo Code}
\label{sec:pseudocode}

This section gives a compact, implementation-faithful view of EPO. We retain only
the routines with a clear semantic role and omit pure book-keeping (timing,
logging, visualization, and I/O) calls. Following \cref{sec:method}, $\mathcal{G}$
denotes the initial geometric estimate $\{(\mathbf{K}_i,\mathbf{P}_i,\mathbf{Z}_i)\}$
produced by the 3DFM and $\mathcal{E}$ the viewgraph of valid pairs. Two algorithms
cover the whole pipeline: construction (\cref{alg:epo-init}, run once) and the
optimization loop (\cref{alg:epo-forward}, run at every iteration). Routines backed by a fused custom Triton kernel are typeset in \T{blue};
everything else is reference PyTorch. The blue routines are the subject of
\cref{sec:triton}, and the stochastic viewgraph sub-sampling inside
\T{\fn{ComputeForwardStep}} is analyzed in \cref{sec:vgsub}.


\begin{algorithm}[t]
\caption{\textsc{EPO.Init} — build the optimization problem (once).
\T{Blue}: fused Triton kernel.}
\label{alg:epo-init}
\begin{algorithmic}[1]
\Require reconstruction path, image folder, depth file; detector, hyper-parameters
\Procedure{Init}{}
  \State $\{\mathbf{K}_i\},\{\mathbf{P}_i\}\gets$ \Call{LoadReconstruction}{path}
        \Comment intrinsics and poses
  \State $\{I_i\}\gets$ \Call{LoadAndPreprocessImages}{folder, size$=518$}
  \State $\{\mathbf{Z}_i\}\gets$ \Call{LoadAndPreprocessDepths}{file}
        \Comment per-pixel depth from the 3DFM
  \State instantiate learnable \textsc{CameraModule}, \textsc{PoseModule}
        \Comment from $\{\mathbf{K}_i\},\{\mathbf{P}_i\}$
  \State $\{\mathbf{E}_i\}\gets$ \Call{ExtractEdges}{$\{I_i\}$; detector}
        \Comment Canny edges, sub-sampled to ${\le}\,E_{\max}$ points / image
  \State $\{\mathrm{DTF}_i\}\gets$ \T{\Call{ComputeDistanceFields}{$\{\mathbf{E}_i\}$}}
        \Comment exact $L_2$ distance transform
  \State $\mathcal{E}\gets$ \Call{BuildViewgraph}{$\{\mathbf{K}_i\},\{\mathbf{P}_i\},\{\mathbf{Z}_i\}$}
        \Comment exhaustive, cycle-consistent 2D--2D reprojection (\cref{sec:viewgraph}; projection in \T{Triton})
  \State instantiate learnable \textsc{DepthModule}
        \Comment per-pixel (scale, shift) over the frozen depth
  \State build per-module optimizers (linear warm-up $\rightarrow$ cosine decay)
\EndProcedure
\end{algorithmic}
\end{algorithm}

\begin{algorithm}[t]
\caption{\textsc{EPO.Forward} — edge-reprojection optimization loop.
\T{Blue}: fused Triton kernel.}
\label{alg:epo-forward}
\begin{algorithmic}[1]
\Require batch size $B$, max iterations $N$, pair budget $M$
\Procedure{Forward}{}
  \State $\mathcal{P}_{\text{prev}}\gets\{\mathbf{P}_i\}$
  \For{$\text{step}=0,\dots,N-1$}
    \State \Call{ZeroGrad}{all optimizers}
    \State \Call{UpdatePoses}{}
          \Comment MLP + $\delta + Gram--Schmidt$
    \State \Call{UpdateIntrinsics}{}
          \Comment per-camera focal scale $f_{\text{eff}}=f\,(1{+}\gamma)$
    \State \T{\Call{UnprojectEdgesTo3D}{}}
          \Comment $\mathbf{X}_i=\mathbf{P}_i^{-1}\,\mathbf{K}_i^{-1}\,[u,v,1]^\top\,\mathbf{Z}_i(u,v)$
    \State $(r,\mathcal{S})\gets$ \T{\Call{ComputeForwardStep}{$\mathcal{E},B,\text{step}$}}
          \Comment \cref{alg:epo-step}
        \State $\ell\gets$ \Call{ComputeBatchedLoss}{$r$}
          \Comment $\sum$ the two directions per pair, mean over pairs ($\mathcal{L}$); plain autograd ops
    \State \Call{Backward}{$\ell$}
          \Comment autograd; the analytical backward of the two \T{blue} kernels above runs here

\State \Call{OptimizerAndSchedulerStep}{}
          \Comment the depth optimizer steps only once $c_{\text{pose}}$ holds (phase 2)
    \State $\theta^{s}_{95}\gets$ \Call{EvaluatePoseChanges}{$\mathcal{P}_{\text{prev}},\{\mathbf{P}_i\}$};\;
          $\mathcal{P}_{\text{prev}}\gets\{\mathbf{P}_i\}$
          \Comment $\theta^{s}_{95}\!=\!\max(\Delta R^{s}_{95},\Delta t^{s}_{95})$: $95$th-pct.\ per-image pose change
    \If{\textbf{not}\ $c_{\text{pose}}$}
      \If{\Call{Converged}{$\theta^{0:s}_{95};\,w_{1},\epsilon_{1}$}}\ $c_{\text{pose}}\gets\textbf{true}$
            \Comment phase 1$\,\to\,$2: enable depth optimizer
      \EndIf
    \ElsIf{\Call{Converged}{$\theta^{0:s}_{95};\,w_{2},\epsilon_{2}$}}\ \textbf{break}
          \Comment phase-2 stop: pose-change criterion
    \EndIf

  \EndFor
\EndProcedure
\end{algorithmic}
\end{algorithm}

\begin{algorithm}[t]
\caption{\textsc{ComputeForwardStep} — batched residuals with stochastic
viewgraph sub-sampling. \T{Blue}: fused Triton kernel.}
\label{alg:epo-step}
\begin{algorithmic}[1]
\Require viewgraph $\mathcal{E}$, batch size $B$, step; pair budget $M$
\Function{ComputeForwardStep}{$\mathcal{E},B,\text{step}$}
  \State $\lambda\gets$ \Call{AnnealClamp}{step}
        \Comment clamp $\lambda:10\!\rightarrow\!6$ over the first $1000$ steps
  \If{$|\mathcal{E}| > M$}
        \Comment stochastic viewgraph sub-sampling (\cref{sec:vgsub})
    \State $\mathcal{S}\gets$ uniform random subset of $\mathcal{E}$, $|\mathcal{S}|=M$
          \Comment fresh draw \emph{each} step
  \Else
    \State $\mathcal{S}\gets\mathcal{E}$
  \EndIf
  \State $r\gets[\,]$
  \For{$b$ in \Call{Split}{$\mathcal{S},B$}}
    \State $(\mathbf{X}_i,\mathbf{K}_j,\mathbf{P}_j,\mathrm{DTF}_j)\gets$ \Call{CreateBatchedInputs}{$b$}
          \Comment $i\!\rightarrow\!j$, $j\!\rightarrow\!i$
    \State $(\rho,\text{valid})\gets$
          \T{\Call{ProjectSampleHuber}{$\mathbf{X}_i,\mathbf{K}_j,\mathbf{P}_j,\mathrm{DTF}_j;\lambda$}}
    \State append per-direction mean $\big(\textstyle\sum_n \rho_n \big/ \sum_n \text{valid}_n\big)$ to $r$
          \Comment reduction shown unfused; benchmarked config fuses it in-kernel (\cref{sec:triton})
  \EndFor
  \State \Return \Call{Concat}{$r$}, $\mathcal{S}$
\EndFunction
\end{algorithmic}
\end{algorithm}

\subsection{Construction and Optimization Loop}
Following the loop order, \T{\fn{UnprojectEdgesTo3D}} performs the inverse mapping
$(\mathbf{u},\mathbf{Z}_i)\mapsto\mathbf{K}_i^{-1}\mapsto\mathbf{P}_i^{-1}$ that
back-projects the edge pixels to world space (the first half of $\pi$);
\T{\fn{ProjectSampleHuber}} then projects the world points $\mathbf{X}_i$ into the
target frame (the second half of $\pi$), bilinearly samples $\mathrm{DTF}_j$, and
applies the clamp-and-Huber robustifier that defines the per-pixel cost
$\mathcal{H}(\hat d_{i\to j})$ of \cref{eq:dtf_ij}. Both expose hand-written
analytical backwards, so \fn{Backward} (otherwise ordinary autograd) runs the
gradients of these two kernels rather than differentiating their expanded operator
graphs. Notice that splitting $\pi$ lets us lift each camera to world coordinates
only once (this step depends solely on the camera's own attributes in $\mathcal{G}$) and
then project it into each of its target views, saving roughly half of the computation otherwise needed.
The convergence test \fn{Converged} of \cref{alg:epo-forward} is windowed: it
smooths the scalar pose-change signal $\theta^{s}_{95}=\max(\Delta R^{s}_{95},\Delta t^{s}_{95})$
with a moving average over a window of $w$ steps and fires only when every smoothed
value in the window is below $\epsilon$. EPO runs two such phases in sequence: the
first, $(w_1,\epsilon_1)$, refines the camera parameters and, on convergence, switches on the
depth optimizer; the second, $(w_2,\epsilon_2)$, then stops the loop once the pose change
stabilizes.

\subsection{Fused Triton kernels}
\label{sec:triton}

EPO routes its four most compute-intensive geometric operations to custom Triton kernels~\cite{tillet2019triton}. In reference
PyTorch, each path expands into a chain of four to six tensor operations, and every
operation incurs a separate CUDA kernel launch, an autograd node, and an
intermediate-tensor allocation. We fuse each chain into a single kernel and supply
a hand-written analytical backward wherever gradients are required. We introduce
the kernels in pipeline order: two run once at construction and are forward-only,
and two run at every iteration and carry analytical backwards.

\paragraph{Construction kernels (run once).}
The distance fields are produced by an exact $L_2$ Euclidean distance
transform~\cite{felzenszwalb2012distance}; evaluating it as a Triton kernel keeps
the field on the GPU and removes the OpenCV CPU round-trip of the reference
implementation. The viewgraph $\mathcal{E}$ is then obtained by the exhaustive,
cycle-consistent 2D--2D reprojection of \cref{sec:viewgraph}: for every one of the
$\binom{N}{2}$ candidate pairs a Triton kernel projects a grid of source pixels into
the target view and back, and pairs whose round-trip-consistent points fall above a
threshold are discarded. Only this forward-and-back projection is the Triton kernel;
the pair enumeration and the consistency/pruning logic around it stay in reference
code.

\paragraph{Per-iteration kernels (run every step).}
Following the loop order, \T{\fn{UnprojectEdgesTo3D}} back-projects each image's (sub-sampled) edge pixels to world space once per iteration. \T{\fn{ProjectSampleHuber}} then projects these points into each target frame and samples its $\mathrm{DTF}$ for every pair. Performing the back-projection once per image rather than once per pair halves the lifting cost, since the source points shared across a pair's reprojections are computed only once. Projection and sampling dominate the optimization cost, as they are evaluated for every pair in the sampled viewgraph. Both kernels fuse the forward pass with an analytical backward.

The fused kernels are faster for three reasons: each multi-operation PyTorch chain,
together with its autograd nodes, collapses into a single launch, removing most
kernel-launch overhead; the operator graph's large intermediates are never
materialized, reducing memory traffic in both the forward and the backward; and the
kernels are scheduled for cache locality on the dominant distance-field sampling.
Each kernel is a numerically equivalent substitute for the corresponding reference
operations---agreeing up to floating-point accumulation order and verified by gradient
and equivalence tests---so the \texttt{triton} backend is effectively a
\textit{drop-in replacement} that preserves the optimization result while reducing
runtime. The two per-iteration kernels are enabled by the \texttt{triton} backend
(the configuration we evaluate), while the reference PyTorch chains remain available;
the distance transform additionally falls back to OpenCV when CUDA/Triton is
unavailable. The project-and-sample kernel additionally offers an optional in-kernel
fusion of the per-pair loss reduction --- the setting we benchmark --- in which each
block reduces its own tile of residuals and only the per-tile partial sums are
accumulated afterwards. This lowers memory traffic, as the full per-edge residual
tensor is never materialized, at the cost of exact bit-reproducibility against the
reference reduction order, since the accumulation order differs (the resulting
difference is zero-mean). Timed as wall-clock per
optimization step on a single RTX~4090 in this evaluated configuration, the fused path runs at a near-constant ${\sim}7.3$\,ms versus ${\sim}28$--$39$\,ms for the reference path --- roughly $4.6\times$ faster on average. Because construction is one-time and
identical for both backends, this per-step speed-up is also the end-to-end
throughput gain over a full optimization run.
\subsection{Stochastic viewgraph sub-sampling}
\label{sec:vgsub}
The viewgraph $\mathcal{E}$ contains all image pairs that survive the 2D--2D
reprojection filter, so its size grows as $\mathcal{O}(N^2)$ and a full-graph loss
evaluation at every iteration ties per-step cost and memory to scene size. In
practice we find this is unnecessary: capping the number of pairs evaluated per
step leaves accuracy unchanged while making each step faster. Concretely, at each
iteration we evaluate the loss on a fresh uniform random subset
$\mathcal{S}\subseteq\mathcal{E}$ of size $\min(|\mathcal{E}|, M)$ (default
$M=1024$) rather than on all of $\mathcal{E}$. This amounts to mini-batch SGD over
the constraint set: each subset gradient is an unbiased estimate of the full-graph
gradient, and since a fresh subset is drawn every step, over a run the optimizer
sees the whole viewgraph in expectation while per-step cost and memory stay
independent of $|\mathcal{E}|$. Empirically, AUC saturates before the full
graph is used, thus raising $M$ beyond $1024$ yields no measurable gain while cost grows
linearly. Scenes much larger than $150$ images may benefit from larger $M$.

\section{Stopping Criteria}
\begin{figure}[t]
    \centering
    \begin{subfigure}{0.95\linewidth}
        \centering
        \includegraphics[width=\linewidth]{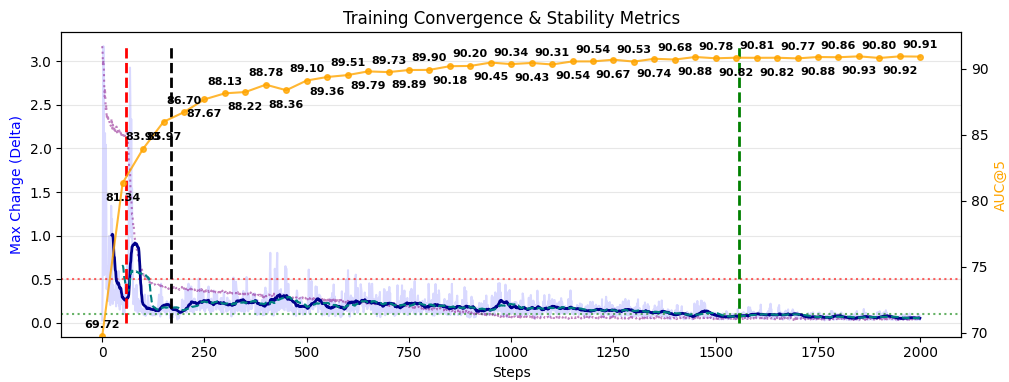}
        \caption{Optimization logs for VGGT+EPO on \textit{Salzburg Rechte Altstadt} (TerraSky3D).}
        \label{fig:stop1}
    \end{subfigure}
    \hfill
    \begin{subfigure}{0.95\linewidth}
        \centering
        \includegraphics[width=\linewidth]{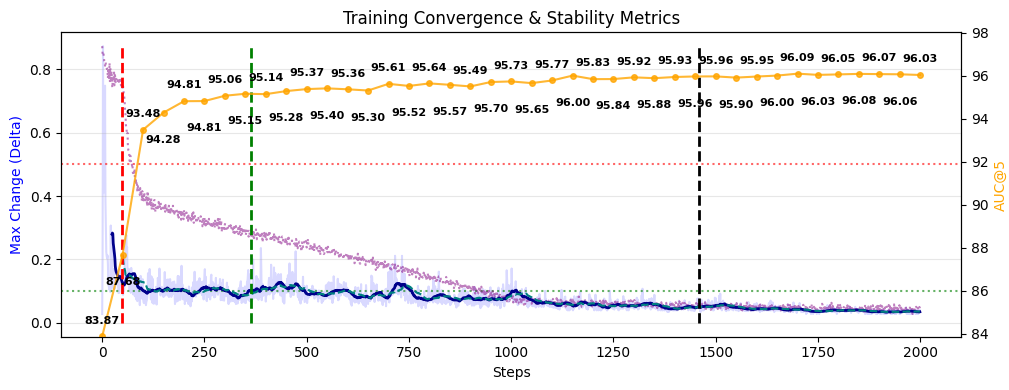}
        \caption{Optimization logs for VGGT+EPO on \textit{Counter} (Mip-Nerf 360).}
        \label{fig:stop2}
    \end{subfigure}
    \caption{The light and dark blue lines represent the maximum pose change and the smoothed maximum pose change, respectively. The dotted-purple and yellow lines represent the loss and AUC. The vertical dashed red line indicates the iteration where the first convergence happens. The green dashed line marks where our \textit{pose-based} stopping criterion terminates the optimization, while the black dashed line indicates the same for the \textit{loss-based} one.}
    \label{fig:combined_stopping}
\end{figure}

To visualize the pose criteria during the optimization, we report \cref{fig:combined_stopping} that traces the loss, the maximum pose change, and the AUC with respect to GT for the \textit{Salzburg Rechte Altstadt} and \textit{Counter} scenes. The AUC tends to plateau as a result of the learning rate decay and the loss convergence, yet the \textit{loss-based} and \textit{pose-based} criteria rarely agree on the optimal termination point. In \cref{fig:stop1}, the \textit{loss-based} criterion (vertical black line) triggers termination prematurely, despite the fact that continuing the optimization would be beneficial. Conversely, in \cref{fig:stop2}, it terminates significantly later than the \textit{pose-based} (vertical green line) one while yielding only marginal gains in pose accuracy. By monitoring the pose directly rather than using the loss as a proxy for pose convergence, our \textit{pose-based} criterion stops precisely when the poses stabilize, making it preferable in both regimes. 

\section{NVS Examples}

Additional NVS results on Mip-NeRF 360 \cite{barron2022mip}, following the format of \cref{fig:nvs_render}, are reported below. The sequences illustrate 3DGS renderings starting from (a) the raw VGGT output, from (b) the reconstruction after track refinement and BA (VGGT+Ref+BA), and from (c) our optimized reconstruction (VGGT+EPO). Notably, our approach preserves sharper structural details than BA-based refinement. This stems from the more accurate camera poses and the higher 3D point density provided by our method, which is beneficial for 3DGS optimization. In fact, differently from BA-based methods, where the number of 3D points depends on the number of feature tracks, in our case the number of 3D points depends on the number of pixels in the selected edges. Therefore, we cap the point budget at 100,000 for both VGGT and VGGT+EPO. 

\begin{figure*}[htbp!] 
     \centering
     \begin{subfigure}[b]{0.325\textwidth}
          \centering
          \includegraphics[width=\textwidth]{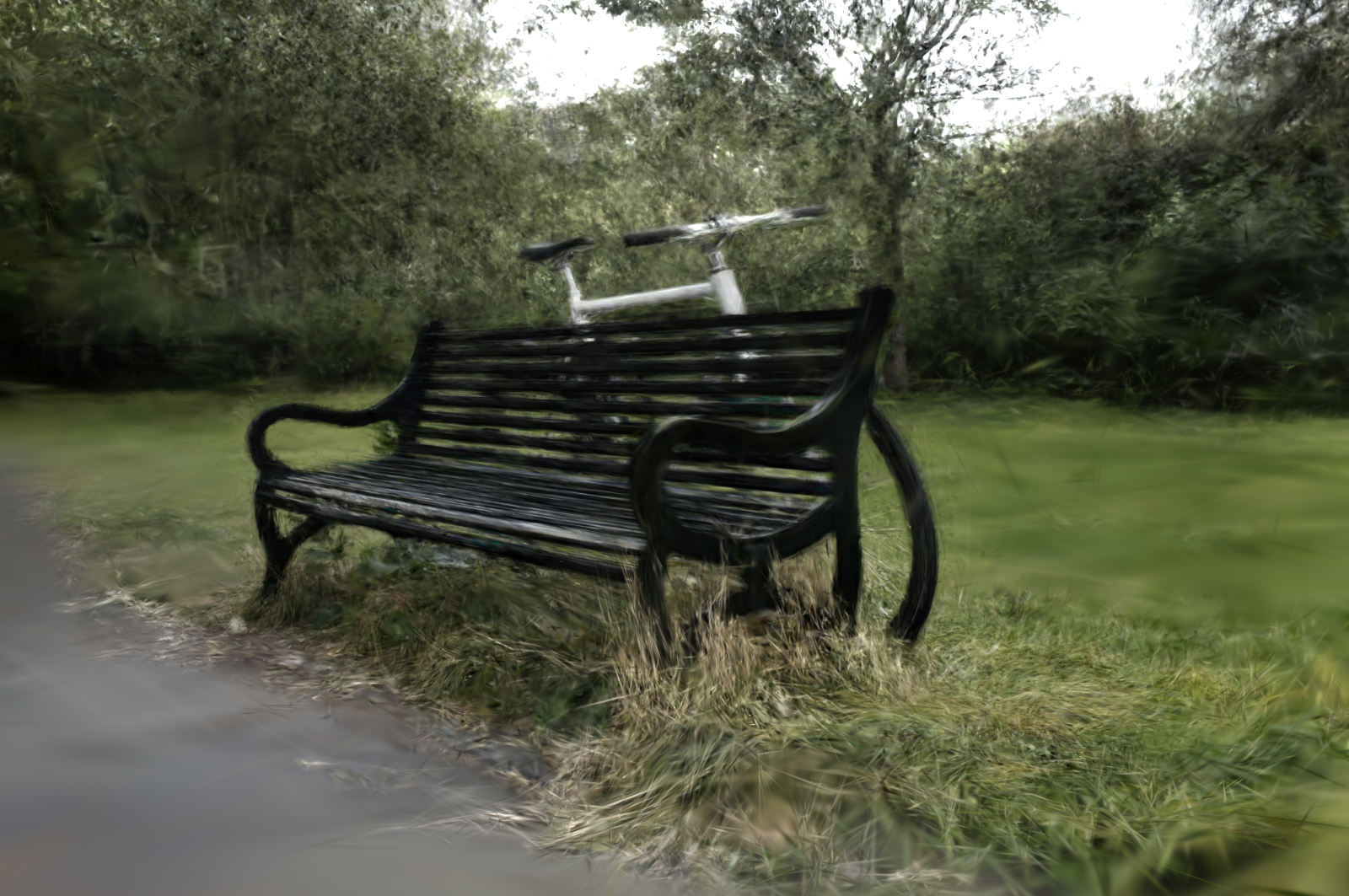}
          \caption{VGGT}
     \end{subfigure}
     \hfill
     \begin{subfigure}[b]{0.325\textwidth}
          \centering
          \includegraphics[width=\textwidth]{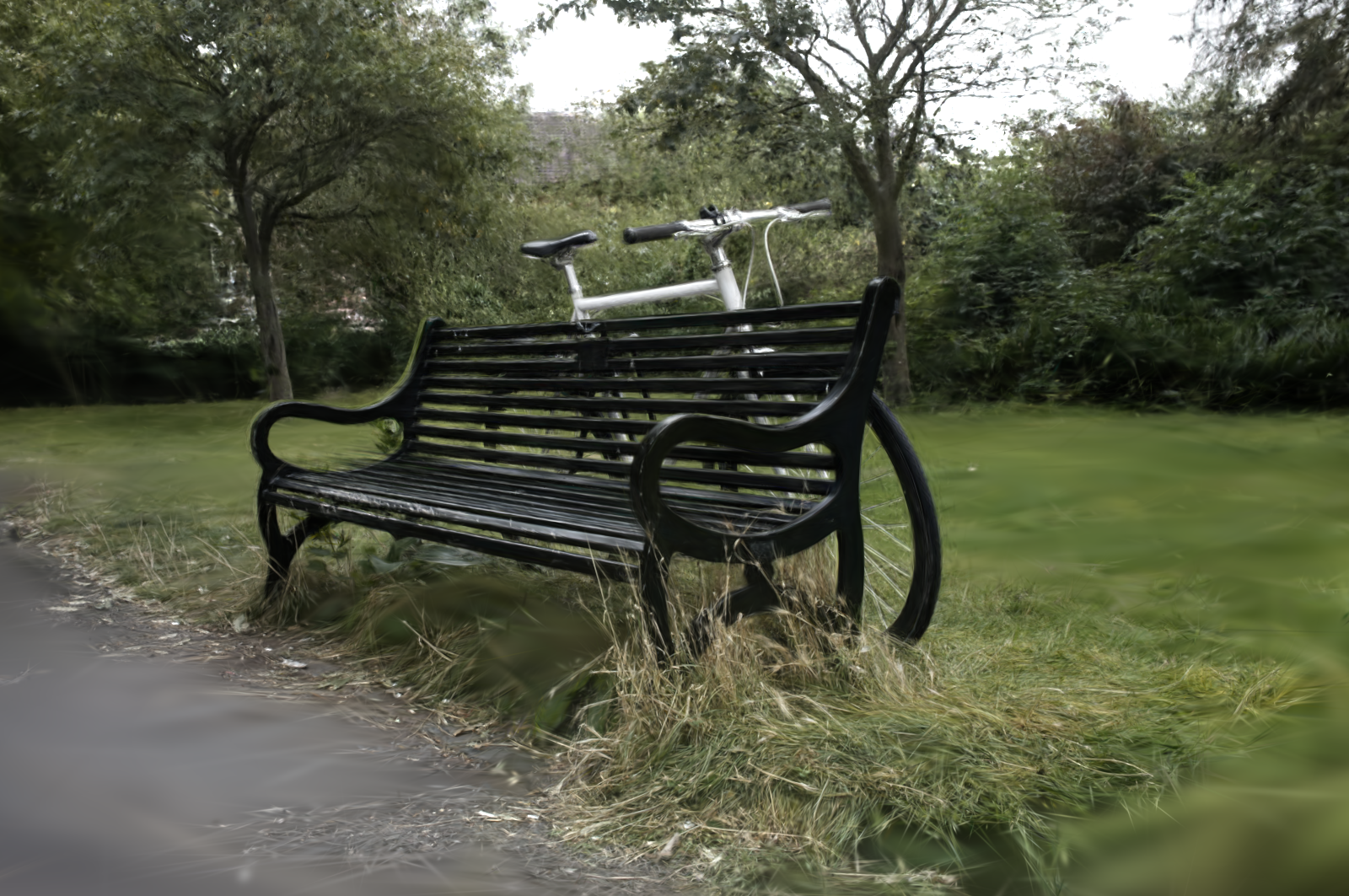}
          \caption{VGGT+Ref+BA}
     \end{subfigure}
     \hfill
     \begin{subfigure}[b]{0.325\textwidth}
          \centering
          \includegraphics[width=\textwidth]{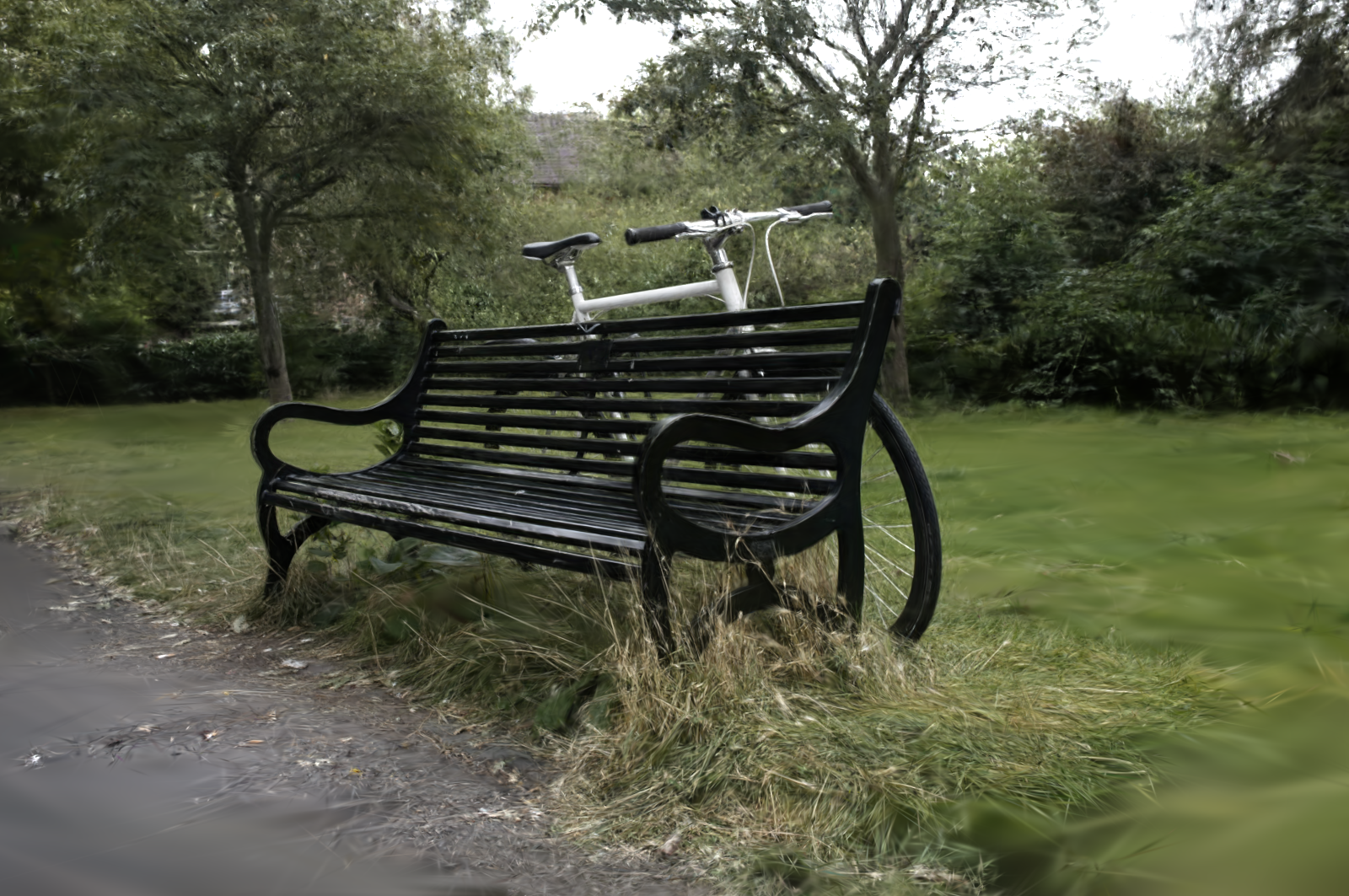}
          \caption{VGGT+EPO (Ours)}
     \end{subfigure}
     \caption{Example renderings from \textit{Bicycle} (Mip-NeRF 360 dataset).}

     \vspace{1.5em} 

     \begin{subfigure}[b]{0.325\textwidth}
          \centering
          \includegraphics[width=\textwidth]{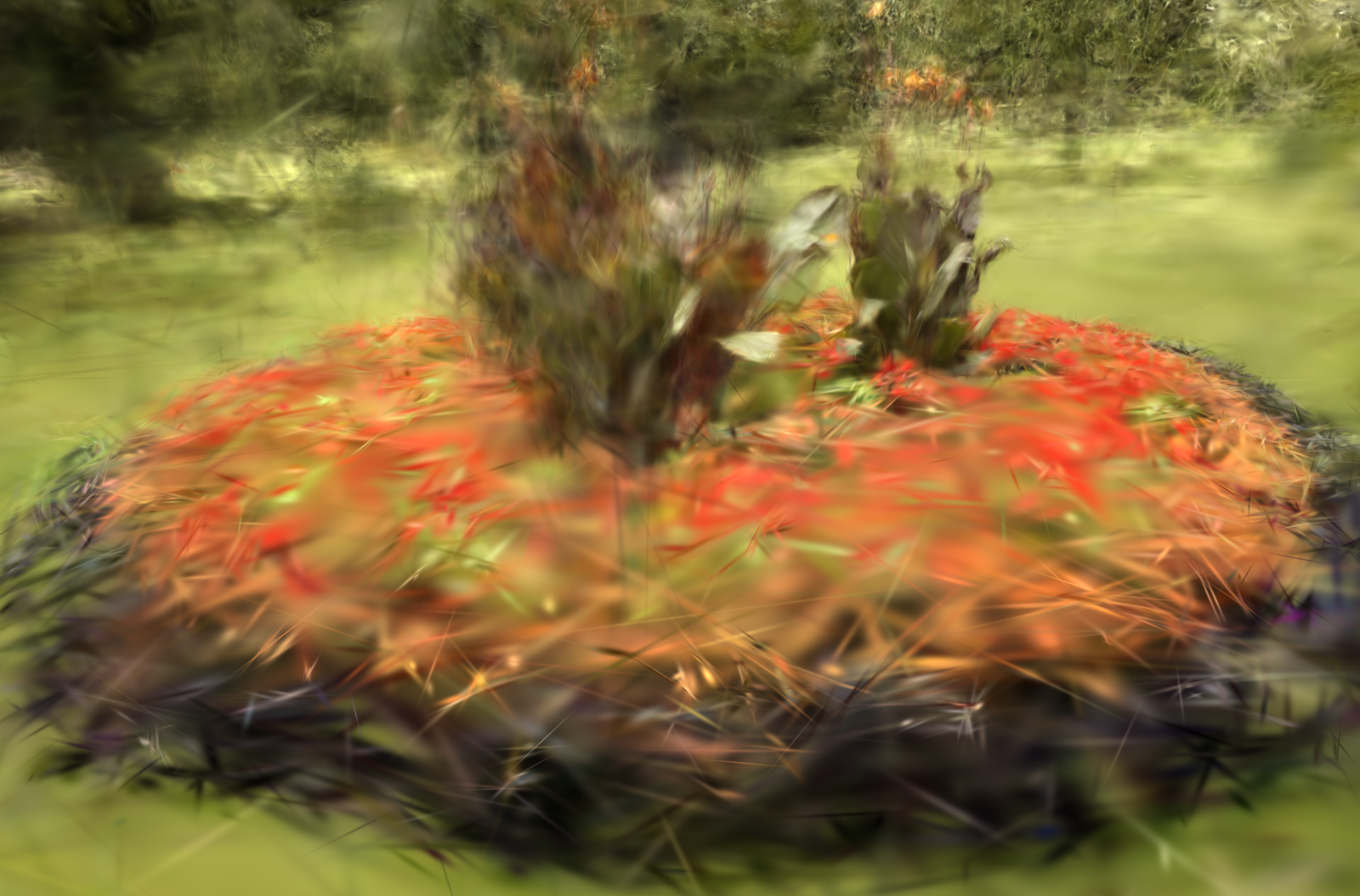}
          \caption{VGGT}
     \end{subfigure}
     \hfill
     \begin{subfigure}[b]{0.325\textwidth}
          \centering
          \includegraphics[width=\textwidth]{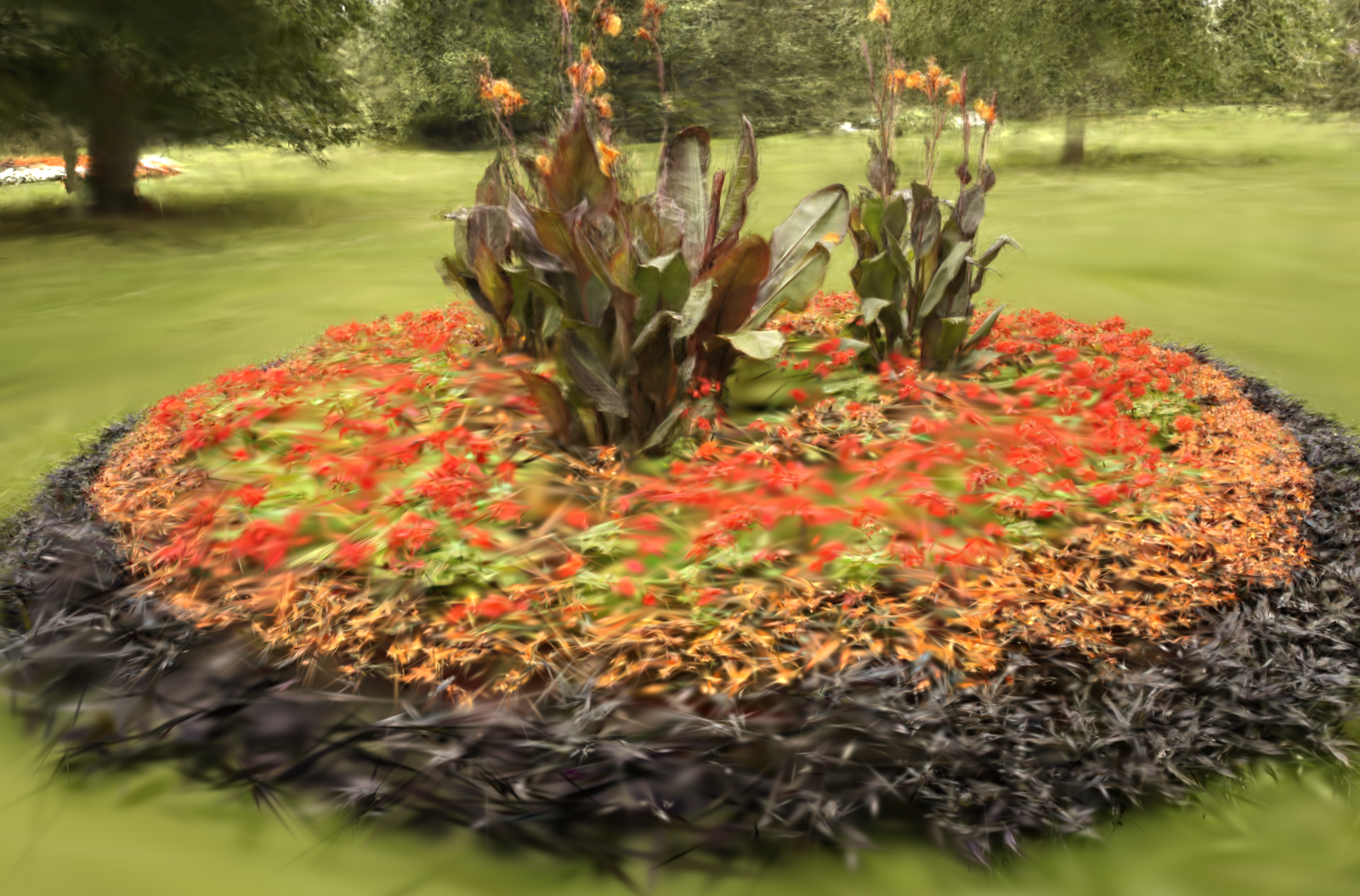}
          \caption{VGGT+Ref+BA}
     \end{subfigure}
     \hfill
     \begin{subfigure}[b]{0.325\textwidth}
          \centering
          \includegraphics[width=\textwidth]{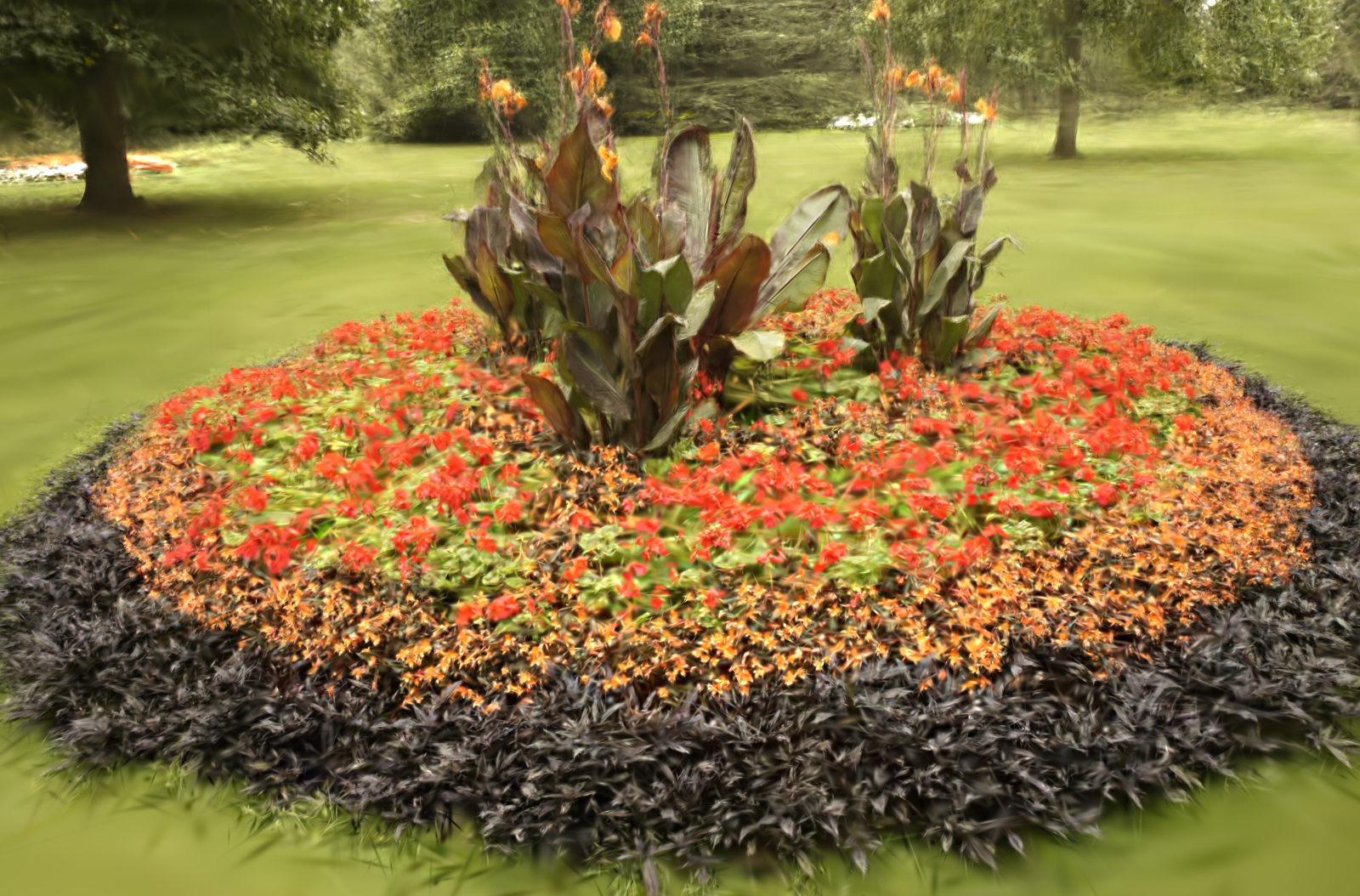}
          \caption{VGGT+EPO (Ours)}
     \end{subfigure}
     \caption{Example renderings from \textit{Flowers} (Mip-NeRF 360 dataset).}

     \vspace{1.5em}

     \begin{subfigure}[b]{0.325\textwidth}
          \centering
          \includegraphics[width=\textwidth]{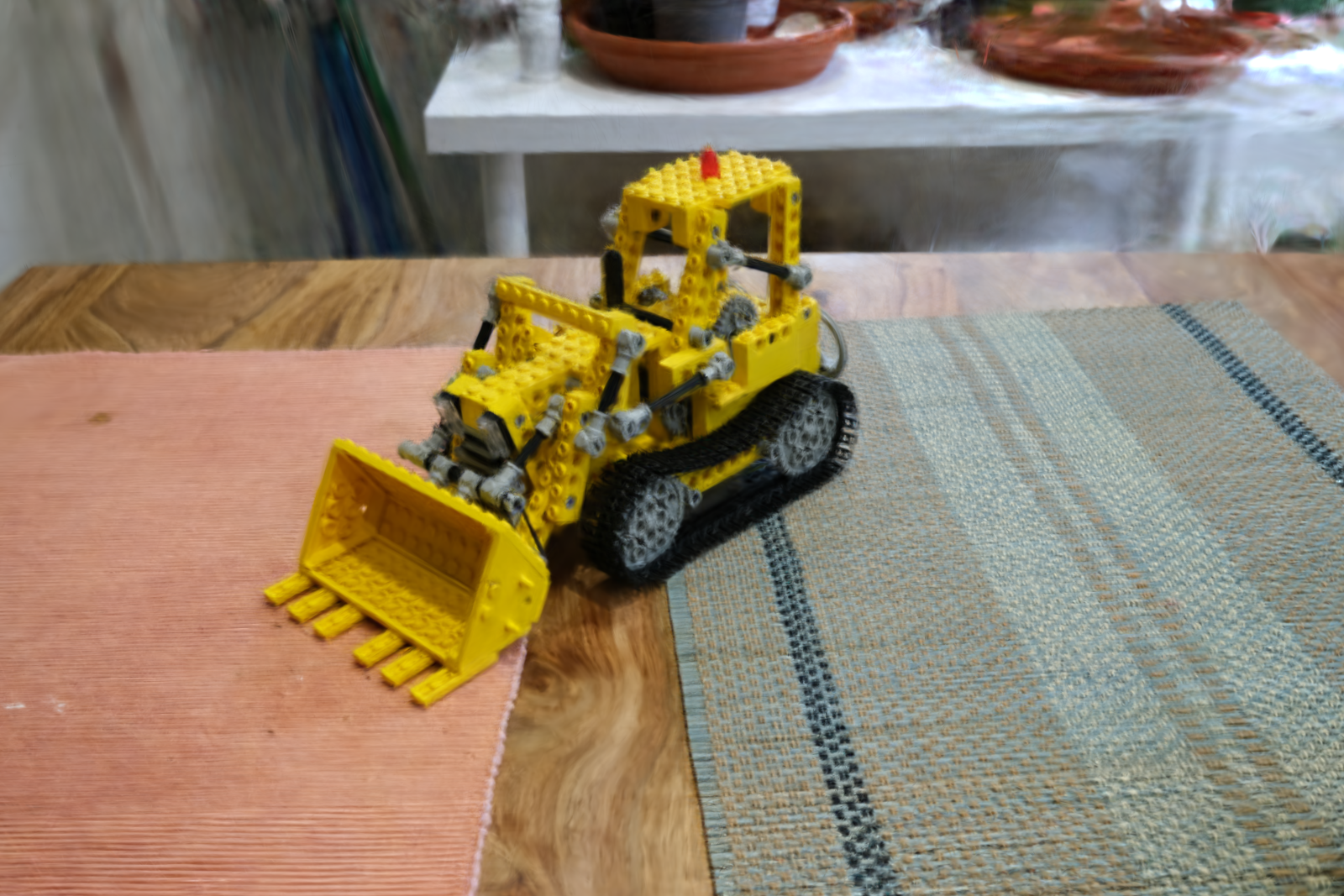}
          \caption{VGGT}
     \end{subfigure}
     \hfill
     \begin{subfigure}[b]{0.325\textwidth}
          \centering
          \includegraphics[width=\textwidth]{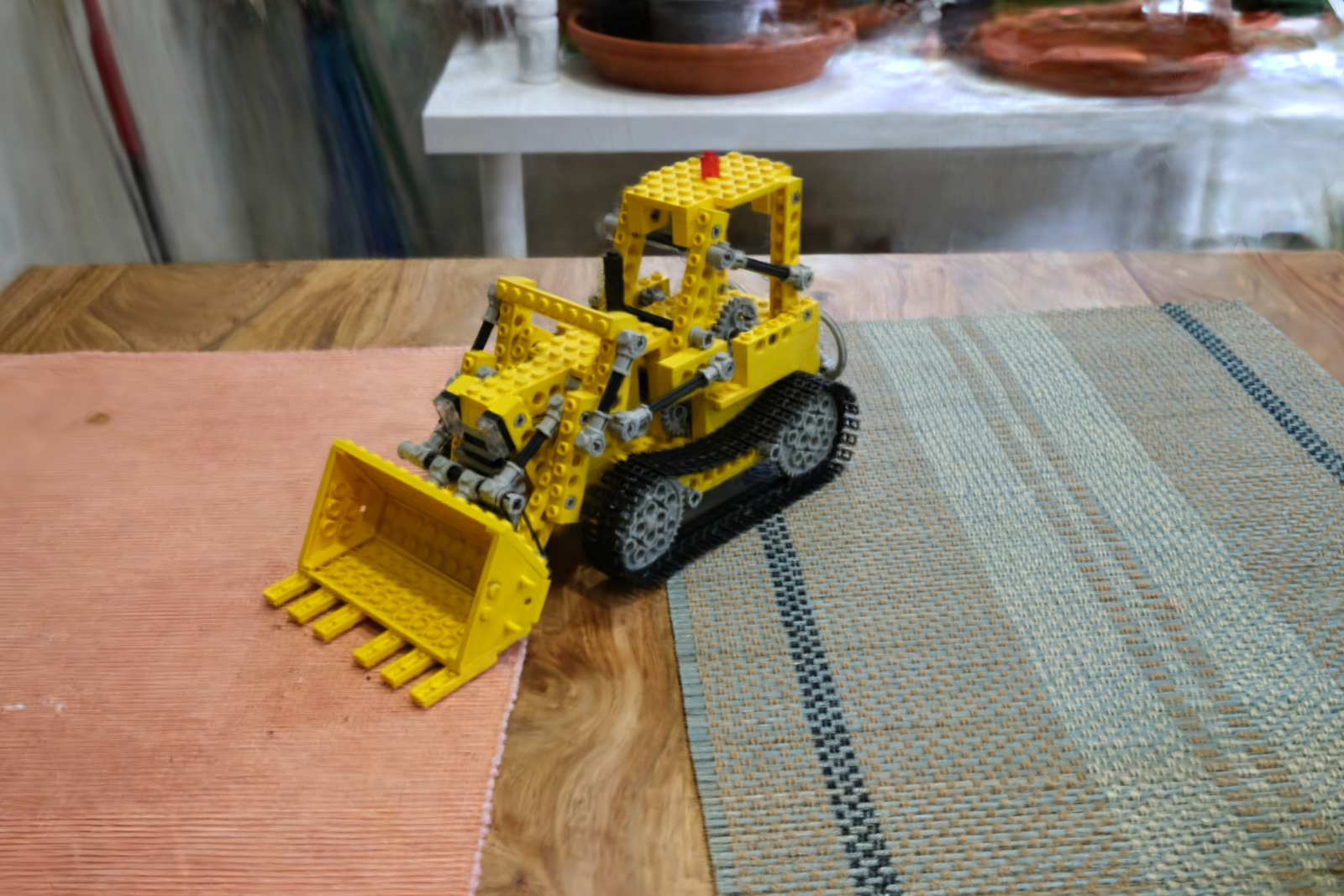}
          \caption{VGGT+Ref+BA}
     \end{subfigure}
     \hfill
     \begin{subfigure}[b]{0.325\textwidth}
          \centering
          \includegraphics[width=\textwidth]{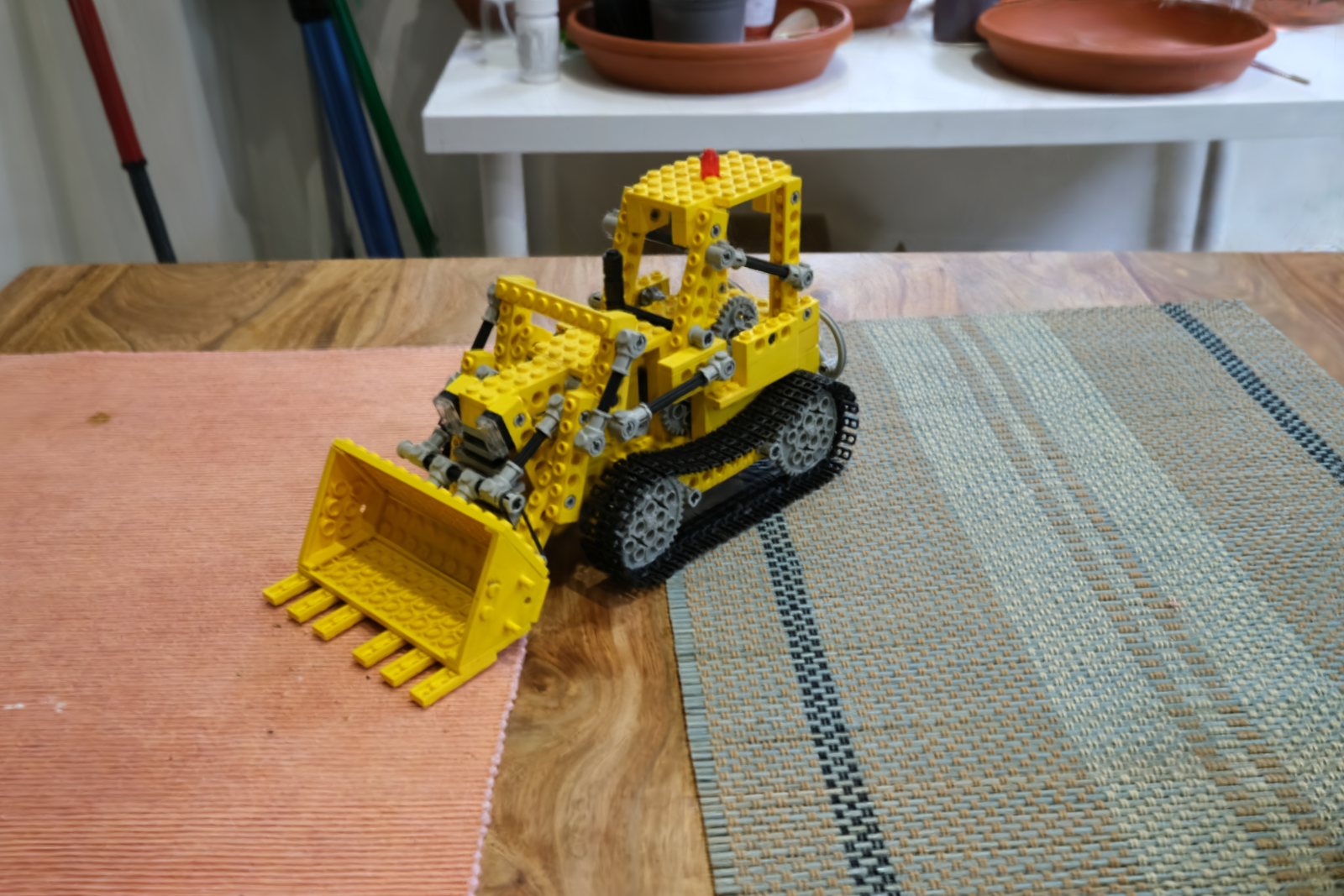}
          \caption{VGGT+EPO (Ours)}
     \end{subfigure}
     \caption{Example renderings from \textit{Kitchen} (Mip-NeRF 360 dataset).}
     
     \label{fig:nvs_render_all}
\end{figure*}
\clearpage

\section{Edge-based 3D Reconstructions}
The following images depict qualitative examples of VGGT 3D reconstructions optimized using EPO's edge-based approach. 

\begin{figure}
    \centering
    \includegraphics[width=0.8\linewidth]{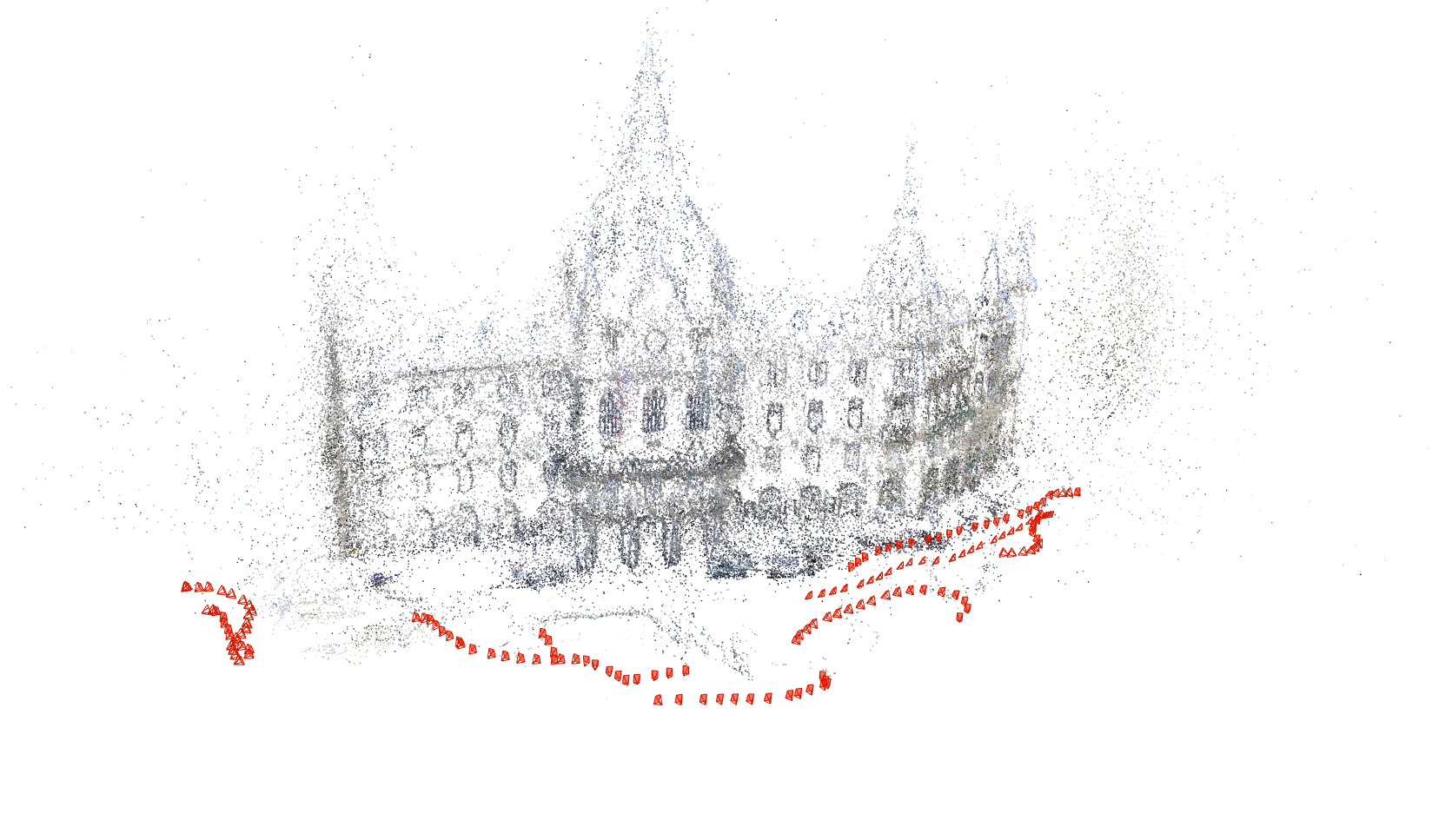}
    \caption{Graz Townhall (TerraSky3D).}
    \label{fig:edge_graz}
\end{figure}

\begin{figure}
    \centering
    \includegraphics[width=0.8\linewidth]{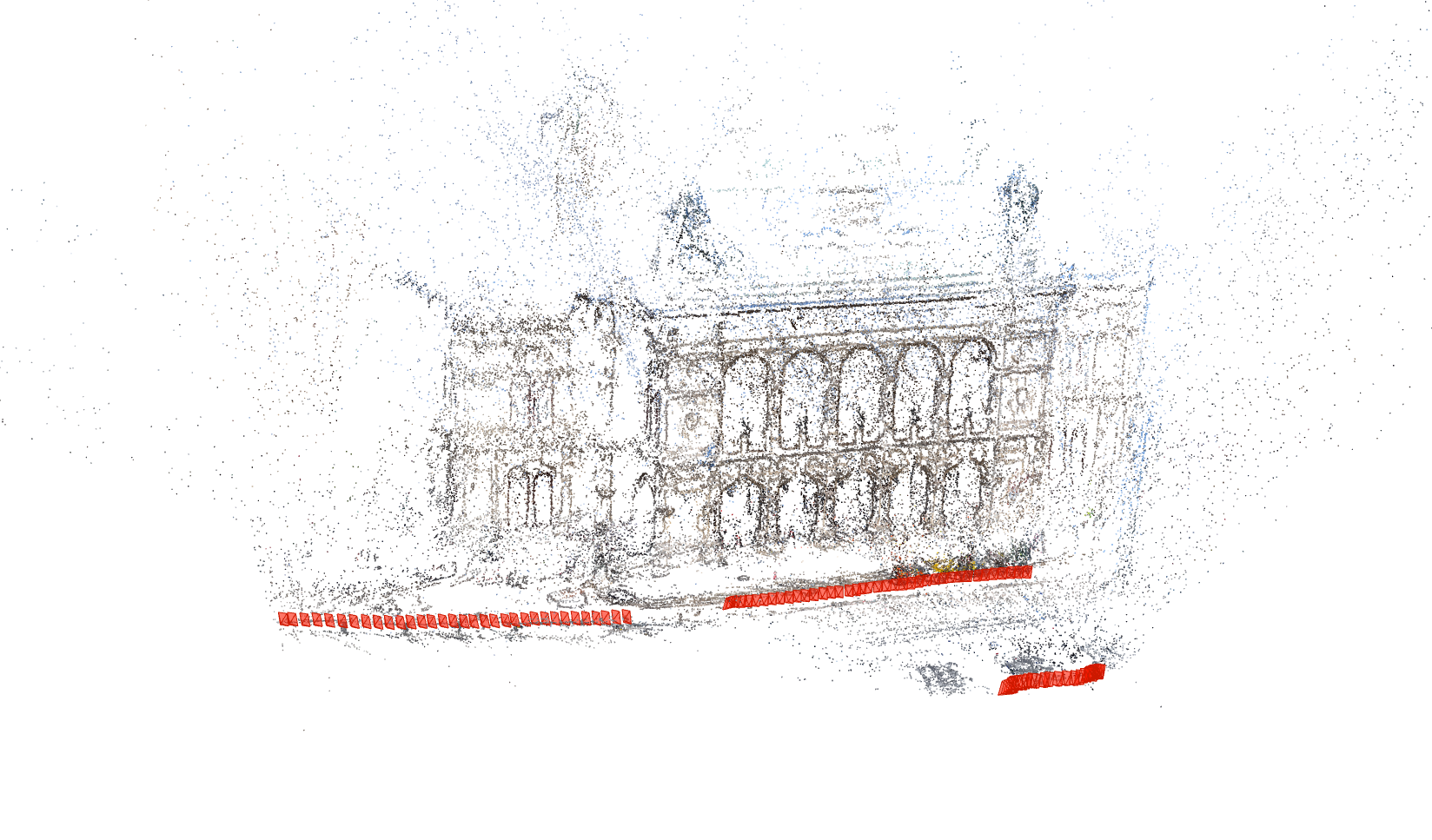}
    \caption{Vienna State Opera (TerraSky3D).}
    \label{fig:edge_vienna}
\end{figure}

\begin{figure}
    \centering
    \includegraphics[width=0.85\linewidth]{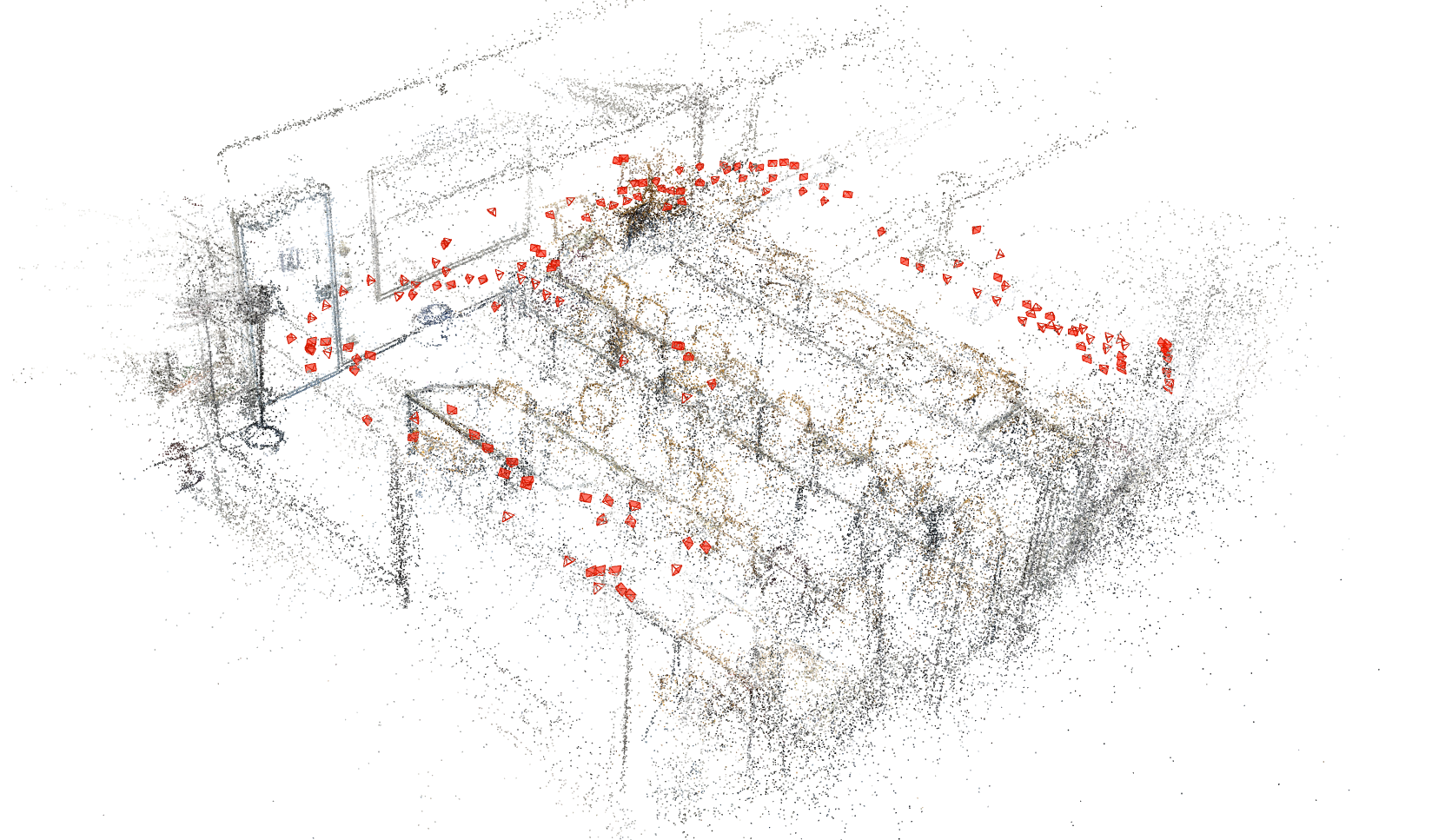}
    \caption{38d58a7a31 (ScanNet++).}
    \label{fig:edge_scannet}
\end{figure}

\begin{figure}
    \centering
    \includegraphics[width=0.85\linewidth]{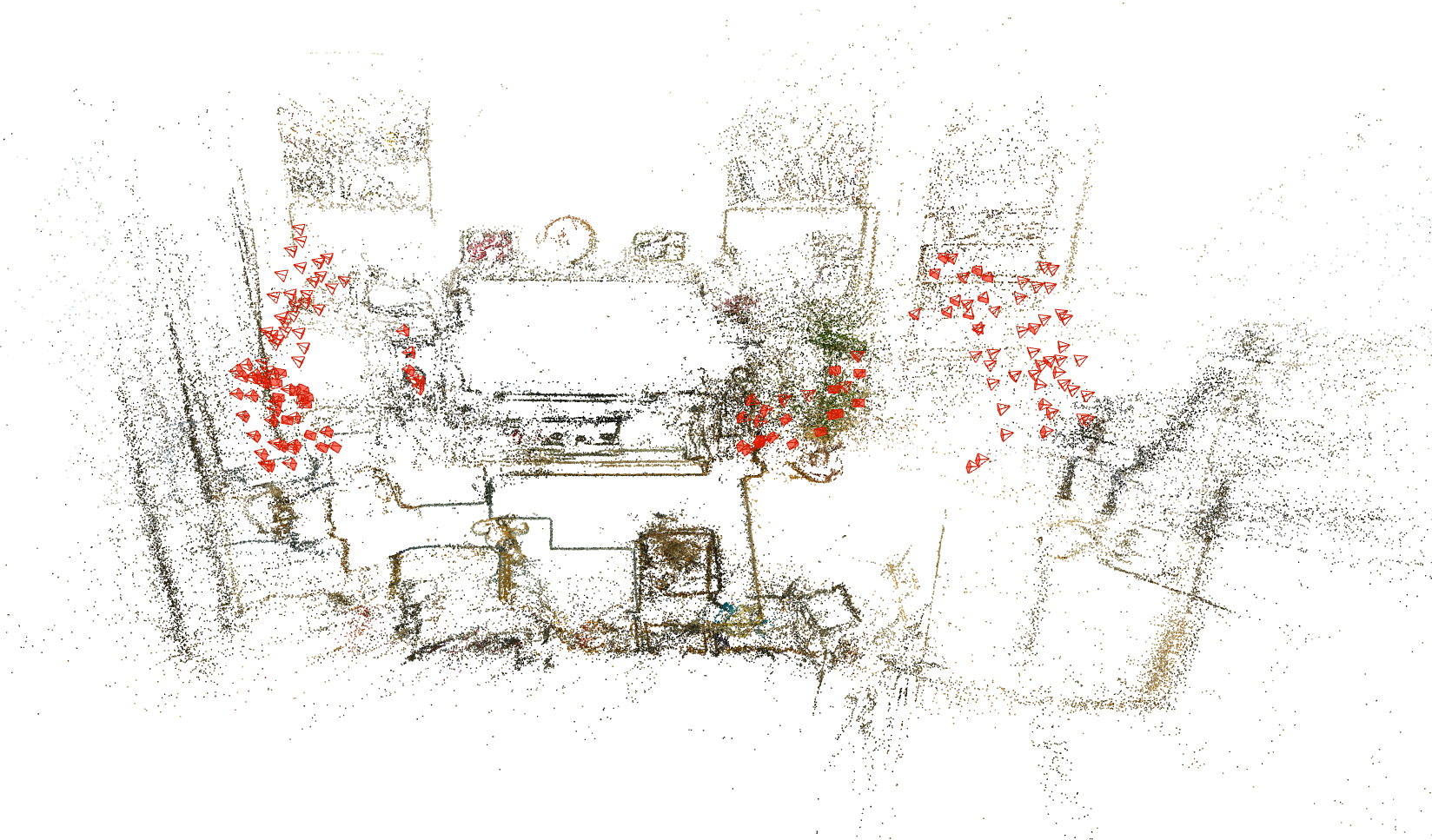}
    \caption{Room (Mip-NeRF 360).}
    \label{fig:edge_room}
\end{figure}

\clearpage


\bibliographystyle{splncs04}
\bibliography{11_references} 


\end{document}